%% file: main.tex
\documentclass[conference]{IEEEtran}
\IEEEoverridecommandlockouts
\usepackage[utf8]{inputenc} % allow utf-8 input
\usepackage[T1]{fontenc}    % use 8-bit T1 fonts
\usepackage{hyperref}       % hyperlinks
\usepackage{url}            % simple URL typesetting
\usepackage{booktabs}       % professional-quality tables

\usepackage{amsmath,amssymb,amsfonts}
\usepackage{graphicx}
\usepackage{textcomp}
\usepackage{xcolor}
\usepackage{nicefrac}       % compact symbols for 1/2, etc.
\usepackage{algorithm}
\usepackage{algorithmic}
\usepackage{mathrsfs}
\usepackage{tabu}
\usepackage{color}
\usepackage{tikz}
\usepackage{wrapfig}
\usepackage{bbm}
\usepackage{multirow}
\usepackage{hhline}
\usepackage{makecell}
\usepackage{xingstyle}
\usepackage{mathtools}
\usepackage{adjustbox}
\usepackage{abraces}
\usepackage[numbers]{natbib}
\usepackage{lipsum}  

\newcommand{\R}{\mathbb{R}}
\newcommand{\M}{\mathcal{M}}

\renewcommand{\L}{\mathcal{L}}

\def\BibTeX{{\rm B\kern-.05em{\sc i\kern-.025em b}\kern-.08em
    T\kern-.1667em\lower.7ex\hbox{E}\kern-.125emX}}
\begin{document}

\title{Dynamic Combination of Heterogeneous Models for \\Hierarchical Time Series}

\author{
\IEEEauthorblockN{Xing Han}
\IEEEauthorblockA{
\textit{UT-Austin}\\
Austin, USA \\
\texttt{aaronhan223@utexas.edu}}
\and
\IEEEauthorblockN{Jing Hu}
\IEEEauthorblockA{\textit{Intuit} \\
Mountain View, USA \\
$\texttt{jing\_hu@intuit.com}$}
\and
\IEEEauthorblockN{Joydeep Ghosh}
\IEEEauthorblockA{\textit{UT-Austin} \\
Austin, USA \\
\texttt{jghosh@utexas.edu}}
}

\maketitle

\begin{abstract}
We introduce a framework to dynamically combine heterogeneous models called \texttt{DYCHEM}, which forecasts a set of time series that are related through an aggregation hierarchy. Different types of forecasting models can be employed as individual ``experts'' so that each model is tailored to the nature of the corresponding time series. \texttt{DYCHEM} learns hierarchical structures during the training stage to help generalize better across all the time series being modeled and also mitigates coherency issues that arise due to constraints imposed by the hierarchy. To improve the reliability of forecasts, we construct quantile estimations based on the point forecasts obtained from combined heterogeneous models. The resulting quantile forecasts are coherent and independent of the choice of forecasting models. We conduct a comprehensive evaluation of both point and quantile forecasts for hierarchical time series (HTS), including public data and user records from a large financial software company. In general, our method is robust, adaptive to datasets with different properties, and highly configurable and efficient for large-scale forecasting pipelines. 
% We provide all implementations of our method in the experiment section.
\end{abstract}
 

\maketitle
\input{tex/intro}
\input{tex/background}
\input{tex/mainmethod}
\input{tex/experiments}
\input{tex/conclusion}

\bibliographystyle{plain}
\bibliography{reference}
% \newpage
\input{tex/appendix}

\end{document}

%% file: tex/intro.tex
\section{Introduction}
\usetikzlibrary{fit,backgrounds} % <-added
%\usetikzlibrary{shadows.blur}
\usetikzlibrary{shapes,arrows}
\usetikzlibrary{positioning}
\tikzstyle{circ} = [draw, circle, fill=white!20, radius=2.6, minimum size=.8cm, inner sep=0pt]
\tikzstyle{line} = [draw]
\tikzstyle{rect} = [draw, rectangle, fill=white!20, text width=2em, text centered, minimum height=2em]
\tikzstyle{longrect} = [draw, rectangle, fill=white!20, text width=6em, text centered, minimum height=2em]
\tikzstyle{line} = [draw, -latex']

Forecasting time-series with hierarchical aggregation constraints is a common problem in many practically important applications \citep{hyndman2011optimal, hyndman2016fast, lauderdale2020model, taieb2017coherent, zhao2016multi}. For example, retail sales and inventory records are normally at different granularities such as product categories, store, city and state \citep{makridakis2020m5, seeger2016bayesian}. Generating forecasts for each aggregation level is necessary in developing both high-level and detailed view of marketing insights. Another prominent example is population forecast at multiple time granularities such as monthly, quarterly, and yearly basis \citep{athanasopoulos2017forecasting}. Normally, data at different aggregation levels possess distinct properties w.r.t. sparsity, noise distribution, sampling frequency etc. A well-generalized forecasting model should not only forecast each time series independently, but also exhibit \textit{coherency} across the hierarchy. However, generating forecasts using regular multivariate forecasting models does not lead to forecasts that satisfy the coherency requirement. 

A major group of works for HTS forecasting employ a two-stage (or post-training) approach \citep{ben2019regularized, hyndman2011optimal, hyndman2016fast, wickramasuriya2015forecasting}, where base forecasts are firstly obtained for each time series followed by a reconciliation among these forecasts. The reconciliation step involves computing a weight matrix $P$ taking into account the hierarchy, which linearly maps the base forecasts to coherent results. This approach guarantees coherent results but relies on strong assumptions (Gaussian error assumption; \citep{hyndman2011optimal, hyndman2016fast, wickramasuriya2015forecasting} also require unbiased base forecast assumption). Moreover, computing $P$ is expensive because of matrix inversion, making this method unsuitable for large-scale forecasting pipelines. Another line of work attempts to learn inter-level relationships during the model training stage. \citep{han2021simultaneously} provides a controllable trade-off between forecasting accuracy of single time series and coherency across the hierarchy, which connects reconciliation with learned parameters of forecasting models. However, one needs to modify the objective functions of the individual models, which is not always possible, especially for encapsulated forecasting APIs that are commonly used in industrial applications. In addition, \citep{taieb2017coherent} proposed to use copula to aggregate bottom-level distributions to higher levels. However, this method only reconciles point forecasts and obtain higher-level distributions in a bottom-up fashion. This causes potential problems such as error accumulation in highly aggregated levels. To avoid this, one needs to require reasonable probabilistic predictions at the bottom-level beforehand. The possible use of high-dimensional copula in real applications is another drawback. \citep{rangapuram2021end} addresses the reconciliation problem during model training stage as well. It first formulates a constrained optimization problem as the objective of reconciliation, and then incorporates this as an add-on layer during model training. However, this method requires distributional assumptions and is specially designed for deep neural networks.

% \textcolor{red}{two additional works...}

\begin{figure*}[t]
\centering
\setlength{\tabcolsep}{1pt} 
\renewcommand{\arraystretch}{1} 
\scalebox{0.94}{
\begin{minipage}{.38\textwidth}
	\begin{tikzpicture}[node distance = 1.cm,auto]
    \node [circ] (step1) {$v_1$};
    \node [circ, below left = 0.3cm and 1.2cm of step1, label={793:{$e_{1, 2}$}}] (step2) {$v_2$};
    \node [circ, below right = 0.3cm and 1.2cm of step1, label={453:{$e_{1, 3}$}}] (step3) {$v_3$};
    \node [circ, below left = 0.3cm and 0.3cm of step2] (step4) {$v_4$};
    \node [circ, below right = 0.3cm and 0.3cm of step2] (step5) {$v_5$};
    \node [circ, below left = 0.3cm and 0.3cm of step3] (step6) {$v_6$};
    \node [circ, below right = 0.3cm and 0.3cm of step3] (step7) {$v_7$};
    \path [line, rounded corners] (step1) -| (step2);
    \path [line, rounded corners] (step1) -| (step3);
    \path [line, rounded corners] (step2) -| (step4);
    \path [line, rounded corners] (step2) -| (step5);
    \path [line, rounded corners] (step3) -| (step6);
    \path [line, rounded corners] (step3) -| (step7);
    \begin{scope}[on background layer]
    	\node [draw,dashed,red,rounded corners,fill=white,fit=(step3)]{};
    \end{scope}
	\draw [line width=1pt] (2.3,-.7) -- (3.3,-1.1) (2.3,-.48) -- (3.3,.5);
    \end{tikzpicture}
\end{minipage}
\hspace{-2.3em}
\begin{minipage}{.62\textwidth}
	\includegraphics[width=\textwidth]{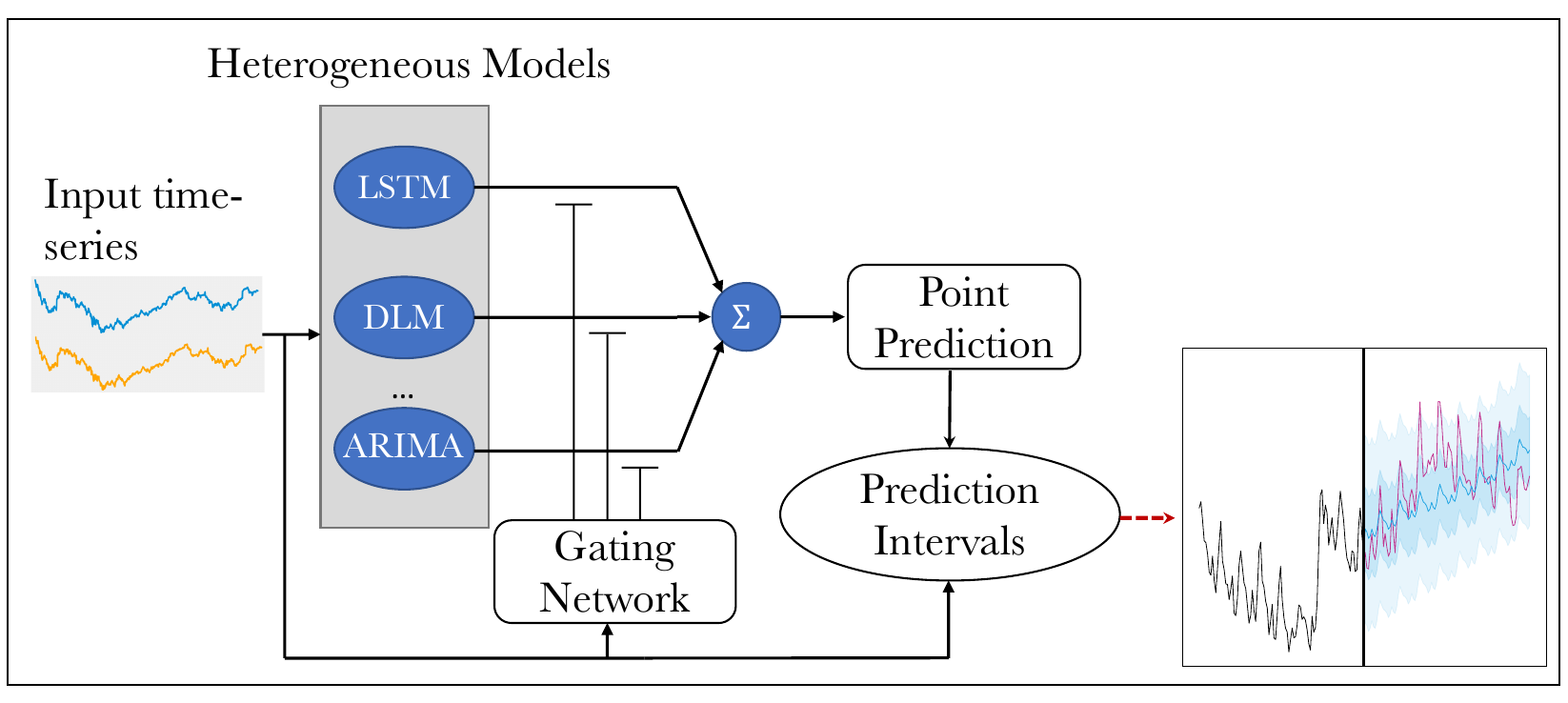}
\end{minipage}}
\caption{Overview of \texttt{DYCHEM}: (left) a three-level structure of HTS (e.g., population forecast at state, city, and county levels); bottom-level vertices: $v_4$ to $v_7$; aggregated-level vertices: $v_1$ to $v_3$; each vertex contains a uni-variate time series. (Right) dynamically combine heterogeneous models; one can mix-and-match different models tailored for each time series.}
\label{fig:overview}
\end{figure*}
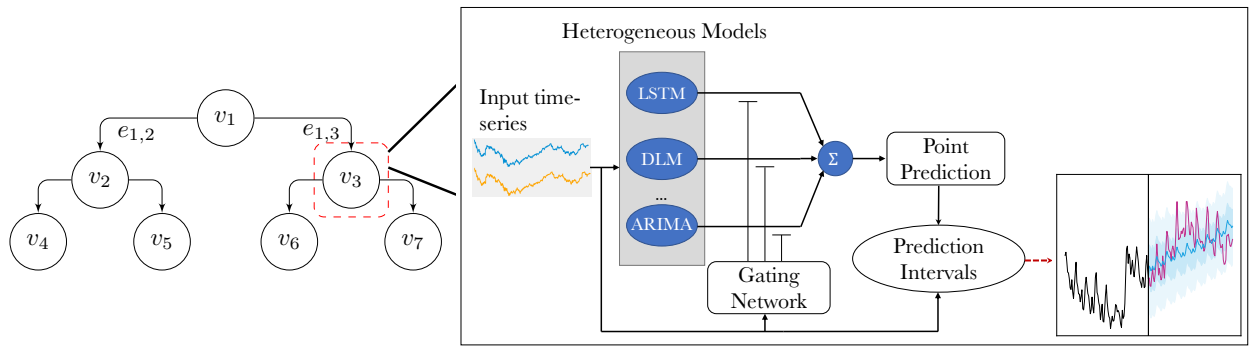

In this work, we propose DYnamic Combination of HEterogeneous Models (\texttt{DYCHEM}) that brings together the power of multiple forecasting models that are inherently different. \texttt{DYCHEM} is primarily designed for a user-specified conceptual hierarchy that contains multiple time series with distinct properties. It can also be applied on regular multivariate or univariate time series. \texttt{DYCHEM} consists of two parts: 1. the point estimator can generate coherent forecasting results and significantly improve the overall accuracy compared with previous baselines; 2. the model-free quantile estimator that takes into a set of quantile levels and simultaneously produce reasonably coherent predictions at these levels. The resulting point estimate can bind the set of obtained quantiles, making each quantile constrained by the hierarchical structure. We show that by diversifying the choice of forecasting models, \texttt{DYCHEM} can provide more accurate forecasts and are therefore more robust to different time series data. Compared with methods using single forecasting model, \texttt{DYCHEM} alleviates the burden of seeking the most appropriate model to a particular time series, since a diverse set of forecasting models is more likely to represent a time series than a single model. This is particularly useful for HTS with multiple aggregated levels, as certain models may specialize in forecasting time series at a particular level. \texttt{DYCHEM} first combines point forecasts from heterogeneous base models, followed by constructing model-free and coherent quantile estimations based on point forecasts. In the following sections, we will show how \texttt{DYCHEM} achieves these properties and provide reliable forecasts in different application scenarios, including a significant improvement on an industrial application in forecasting structured user records.

% talk about Intuit and parallel implementation...

% emphasize the importance of this in industrial applicaitons, no prior work on this, addressing important problems...
% In summary, key contributions of \texttt{DYCHEM} include: 1. improve the performance of point prediction by learning to combine forecasts from a set of heterogeneous models; 2. build user-specified multiple quantile estimations that are model-free and approximately coherent. In addition, \texttt{DYCHEM} can also be applied on general multi-variate or univariate time series data, and is robust to abrupt changes in the data.
% \texttt{MECATS} successfully addresses several key problems in hierarchical time series forecasting, including assigning a suitable model for data with certain properties, enabling an arbitrary forecasting method to learn inter-level relationships, and generating coherent multi-quantile forecasts without quantile crossing. 
% Note that \texttt{MECATS} can also be applied to general sequential data, and we will demonstrate the flexibility of this approach by combining with Bayesian Online Change-point Detection (BOCPD) \citep{adams2007bayesian}.

% add more high-level discussion on current method drawback, the goal to accomplish, advantage of our method

%% file: tex/background.tex
% \vspace{-1.9ex}
\section{Related Works}

\textbf{Combination of Point Forecasts} ~\citep{smith2009simple} has shown that combining different forecasts of the same event by simple averaging is more robust than other methods. However, extra assumption is needed on unbiased base model, which normally fails in some aggregated levels of hierarchical time series. Learning a weight combination by incorporating hierarchical constraints provides a better solution. Mixture of experts (MoE) \citep{jacobs1991adaptive} is a localized, cooperative ensemble learning method for enhancing the performance of machine learning algorithms. It captures a complex model space using a combination of simpler learners. The standard way to learn an MoE is to train a gating network and a set of experts using EM-based methods \citep{yang2009single}. The gating network outputs either experts' weights \citep{chaer1997mixture} or hard labels \citep{ shazeer2017outrageously}. Prior works have studied MoE for regular time series data, where MoE showed success in allocating experts to the most suitable regions of input \citep{lu2006regularized, weigend1995nonlinear}. Hierarchical MoE has also been applied in the speech-processing literature \citep{chen1995speaker} for text dependent speaker identification. However, these works only employed experts of the same type with limited representation power \citep{chaer1997mixture, weigend1995nonlinear}. Most recently, \citep{bhatnagar2021merlion} developed a library to forecast time series using heterogeneous models. The way they combine multiple predictions is through simple averaging or model selection based on a user defined metric. \texttt{DYCHEM} combines point forecasts in a different way with standard MoE. First, the forecasting models are heterogeneous, where each model can be arbitrarily user-specified. Second, it finds better local optimums than EM algorithm by using pre-training warm-up for each model. We will show in the following sections that these properties make \texttt{DYCHEM} more suitable for industrial applications, which leads to improved performance.

\textbf{Uncertainty Estimation} ~The most widely used representation of uncertainty is a forecast interval, which includes information from a marginal distribution: a common choice of parametric distribution is Gaussian. Bayesian methods make assumptions on prior and loss functions \citep{blundell2015weight, iwata2017improving, sun2019functional}. However, these assumptions are not always fulfilled. Multiple observation noise models \citep{salinas2020deepar} and normalizing flow \citep{durkan2019neural, gopal2021normalizing} can generalize to any noise distribution, but it is left to human's expertise to choose appropriate likelihood function. Quantile regression \citep{ han2021simultaneously, tagasovska2018single} avoids distributional assumptions by directly estimating quantiles of distribution via statistical learning methods. Quantile loss can be flexibly integrated with many forecasting models. But extra efforts are needed to prevent quantile crossing, and (possibly) make each quantile coherent across the hierarchical structure. 

% \textcolor{red}{our method is different from MoE in the following points...}
% In this paper, we propose the \texttt{DYCHEM} framework that can not only learn the most suitable adaptive combination of a set of heterogeneous models, but also estimate arbitrary quantiles given the point prediction. \texttt{DYCHEM} is designed for multi-variate time series with hierarchical constraint, and its simplified variant also works for normal time series.

% In this paper, we investigate a more general situation, where the data is not only non-stationary, but multivariate with aggregation constraints. We use \texttt{MECATS} to bring together a heterogeneous set of prominent forecasting models and show that this combination can successfully handle both change of dynamics and hierarchical aggregation.
% In this paper, we adopt gating networks that output experts' weights to combine forecasts from each model.
% we will show when ts changes their dynamics
%. Although DNN models demonstrate prominent advantages as the volume of data grows, they cannot be as flexible and effective as classical methods in any situation. We combine forecasting models from different categories as the set of experts,

%% file: tex/mainmethod.tex
\section{Forecasting HTS using DYCHEM}
\textbf{Problem Setting}
~Figure \ref{fig:overview} (left) shows a hierarchical graph structure with three levels. Each vertex represents time series data aggregated on different variables related through a domain-specific conceptual hierarchy (e.g., product categories, time granularities etc). We use $\{V, E\}$ to represent the graph structure, where $V := \{v_1, v_2, \dots, v_7\}$ is the set of vertices and $E := \{e_{1,2}, e_{1,3}, \dots, e_{3,7}\}$ is the set of edges. Let $x^{v_i}_{1:T}$ be the time series at the vertex $v_i$, where $T$ is the forecast starting point. We will use $x_{1:T}$ instead if the hierarchical information is not important. For the most common use case, we assume $e_{i,j} \in \{-1, 1\}$, and $x^{v_i}_{1:T} = \sum_{e_{i,j} \in E} e_{i,j} ~ x^{v_j}_{1:T}$, which means time series at parent vertices is the (signed) summation of data in their children vertices. Ideally, we want to obtain probabilistic forecasts from $T+1$ to $T+h$:
\begin{equation}
P(\hat{x}^{v_i}_{T+1: T+h}~|~x^{v_i}_{1:T}; \theta), \quad \mathrm{s.t.} ~
| \hat{x}^{v_i}_t - \sum_{e_{i,j} \in E} e_{i,j} ~ \hat{x}^{v_j}_t | = 0, \nonumber
\end{equation}
for any $t \in [T + 1, T + h]$, where $h$ is the forecasting horizon and $\theta$ represents the model's parameters. We define $\delta = \hat{x}^{v_i}_t - \sum_{e_{i,j} \in E} e_{i,j} ~ \hat{x}^{v_j}_t $ as the coherency loss at time $t$ of the hierarchical forecast, which will later be used for evaluations. Note that it is straightforward to extend linear aggregations to non-linear case and weighted edges. The $\{V, E\}$ representation is equivalent to the $S$ matrix used in the most recent hierarchical forecasting literature \citep{ben2019regularized, hyndman2011optimal, hyndman2016fast, wickramasuriya2015forecasting}.

% \textcolor{red}{coherency loss here}

\subsection{Dynamic Combination of Point Forecasts}
Given a model set $\M = \{\mathsf{M_1}, \dots, \mathsf{M_L}\}$ that contains $\mathsf{L}$ base models parameterized by $\{\Theta_1, \dots, \Theta_{\mathsf{L}}\}$, and a gating network $\Theta_g$. Let $\Theta = \{\Theta_g, \Theta_1, \dots, \Theta_{\mathsf{L}}\}$ be the collection of parameters, the point (mean) forecast from $T+1$ to $T+h$ is
\begin{align}
    \mathbb{E}[\hat{x}_{T+1: T+h}~|~x_{1: T}, \Theta] = \sum_{l=1}^{\mathsf{L}} w_l \cdot \mathbb{E}[\hat{x}_{T+1: T+h}~|~ \varphi_l(x_{1: T}), \Theta_l], \nonumber
    % \label{eq:moe}
\end{align}
where $w_l = P(l~|~x_{1: T}, \Theta_g) \in [0, 1]$ is the $l^{th}$ output of gating network, 
% henceforth denoted by $g_l(x^{v_i}_{1: T_i}, \Theta_g)$ instead. 
and $\varphi_l$ is the data preprocessing function of the $l^{th}$ model. 
% Since the experts within $\M$ may have distinct ways to represent predictive uncertainty, it is unreasonable to simply combine their prediction intervals. We therefore restrict our discussion to point prediction $\mathbb{E}[\hat{x}^{v_i}_{T_i+1: T_i+h}~|~l, x^{v_i}_{1: T_i}, \Theta_e]$ from each expert for now.
% , i.e., $P(x^{v_i}_{T_i+1: T_i+h}~|~l, x^{v_i}_{1: T_i}, \Theta_e)$ in Eq (\ref{eq:moe}) is a delta distribution.
EM algorithm is a commonly used method to train MoE, where the latent variable and model parameters are iteratively estimated. However, the EM training is likely to cause load imbalance, which is a self-reinforced procedure that stronger experts get more training \citep{shazeer2017outrageously}. We then disentangle the training of base models and gating network by pre-training each base model offline: this waives the $M$-step of EM algorithm, which simplifies the procedure if each base model has different representation of the input data. In practice, this pre-training warm up finds better local optimum than EM algorithm. For $t_0 \in [1, T]$, each forecasting model is pre-trained on $x_{1:t_0}$ and generate point forecast $\{\hat{x}_{t_0: T}(l)\}_{l=1}^{\mathsf{L}}$. 
% by having the set of experts $\{\mathsf{M}_l\}_{l=1}^{\mathsf{L}}$ trained offline on $x_{1:t_i}^{v_i}$, we first obtain the set of pre-trained experts that generate point predictions $\{\hat{x}^{v_i}_{t_i+1: T_i}(l)\}_{l=1}^{\mathsf{L}}$. 
The training time series is then processed by a sliding window that is used to train the gating network $\mathsf{NN}_g$: $$\mathrm{sw}(x_{1:t_0}) = [x_{1:\omega}, x_{2:\omega+1}, \dots, x_{t_0 - \omega + 1:t_0}] \in \R^{(t_0 - \omega + 1)\times \omega \times 1},$$
% The gating network is then trained on the same data using the sliding window representation. Specifically,
% \begin{equation}
% \mathrm{sw}(x_{1:t_i}^{v_i}) = [x_{1:\omega}^{v_i}, x_{2:\omega+1}^{v_i}, \dots, x_{t_i - \omega + 1:t_i}^{v_i}] \in \R^{(t_i - \omega + 1)\times \omega \times 1}, \nonumber
% \end{equation}
where $\omega$ is the window length, and $\{w_l\}_{l=1}^{\mathsf{L}} = \mathsf{NN}_g(\mathrm{sw}(x_{1:t_0}))$ is the set of weights produced by gating network. The gating network consists of an LSTM recurrent layer followed by a dense layer with ReLU activation and the Softmax function.

% and one can train the gating network efficiently by maximizing the likelihood of weighted combination of each expert.
% we bring together both high variance and low variance experts and disentangle the expert's training with gating network training, i.e., train gating separately with gradient descent.
% high variance & low variance model, formulate ME learning process
% Divide training and validation set, generalization
% gating network structure, sliding window representations

% \subsection{Reconciling Gating Network during Training}
% \textbf{Gating Network} ~ structure of gating network is ...

\begin{figure*}[t]
    \centering
    \includegraphics[width=.99\linewidth]{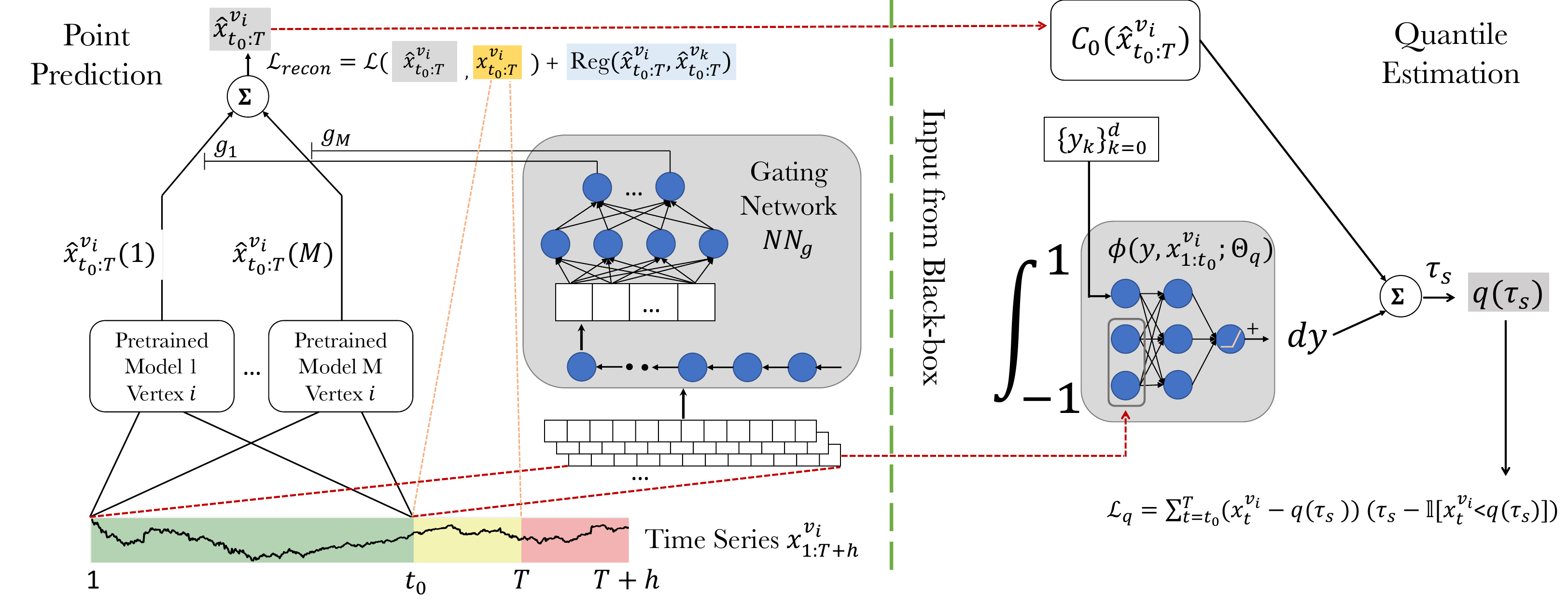}
    \caption{Model structure for generating point and quantile forecasts at vertex $i$. Left: point prediction by combining heterogeneous models, $\L_{\mathrm{recon}}$ is used to train gating network. Right: quantile estimator built on top of black-box point prediction.}
    \label{fig:gating}
\end{figure*}

We then train the gating network on a separate training set from $t_0$ to $T$ that all base models don't have access to. The outputs $\{w_l\}_{l=1}^{\mathsf{L}}$ captures the generalization ability of each pre-trained model, i.e., higher $w_l$ indicates higher importance of the $l^{th}$ model for a particular vertex. The objective for training the gating network at vertex $v_i$ is
\begin{equation}
    \L_{\mathrm{recon}} = \L_{v_i}(\hat{x}^{v_i}_{t_0: T}, x^{v_i}_{t_0: T}) + \lambda \cdot \|\hat{x}^{v_i}_{t_0: T} - \sum_{e_{i,j} \in E} e_{i,j} ~ \hat{x}^{v_j}_{t_0: T}\|^2,
    \label{eq:loss}
\end{equation}
where $\hat{x}^{v_i}_{t_0: T} = \sum_{l=1}^{\mathsf{L}}w_l~\hat{x}^{v_i}_{t_0: T}(l)$ contains gating network parameters $\Theta_g$ at $v_i$. The loss $\L_{\mathrm{recon}}$ not only involves forecast at $v_i$, but also forecasts at its child vertices. By minimizing $\L_{\mathrm{recon}}$, we enable the gating network at $v_i$ to learn across adjacent levels, and maximize the likelihood of $x^{v_i}_{t_0: T}$ through controlling the weight on each model. In addition, we bypass adding regularization terms for each model as \citep{han2021simultaneously} did, which is often impractical. The overall structure of the point forecasting framework is shown in Figure \ref{fig:gating} (left).

Eq (\ref{eq:loss}) provides a controllable trade-off between coherency and accuracy at $v_i$, which helps the forecasting model generalize better, and is more effective than two-stage reconciliation methods when applying on large-scale hierarchical structures. This formulation can also be combined with a \textit{bottom-up training} approach, where the gating networks at the bottom level are first trained without regularization terms, and use the bottom-level forecasting results to progressively reconcile higher-level gating networks using Eq (\ref{eq:loss}), till the root is reached. In contrast, one needs to reconcile both higher (previously visited) and lower-level model at an intermediate vertex if the top-down training method is applied, since other forecasts at that intermediate level might have changed. This bottom-up training procedure can be run in parallel on training vertices at the same aggregation level because they are mutually independent. Therefore, one can efficiently obtain coherent forecasts through one pass of the bottom-up training. 
% We show that by minimizing Eq (\ref{eq:loss}), 
% This is because time series at lower aggregation level
% It is important that the set $x^{v_i}_{t_i+1: T_i}$ will be reserved in model training. 
% A generalization bound might be helpful...

We now discuss how \texttt{DYCHEM} provide accurate and coherent forecasts and how it is different from other methods such as SHARQ \citep{han2021simultaneously} and MinT \citep{hyndman2011optimal}.

% \begin{pro}[Coherency Condition of DYCHEM] \label{thm:coherency}
% Assume $k$ models are assigned to $v_1$, and its corresponding child vertices $\{v_i\}_{i=2}^n$. Each expert generates forecast $\{\hat{x}_{v_i}^j\}_{j=1}^k$, which are not necessarily unbiased.
% \end{pro}
\textbf{Proposition 1} [\textit{Coherency of \texttt{DYCHEM}}] \label{prop:coherency}
\textit{Given $\mathsf{L}$ models are assigned to $v_1$ and its child vertices $\{v_i\}_{i=2}^n$. The models generate point forecast $\{\hat{x}_{v_i}^l\}_{l=1}^{\mathsf{L}}$ at $v_i$, where $\hat{x}_{v_i}^l \in \R$ is not necessarily unbiased. For all $\mathsf{L}$ models, assume their coherency loss $\{\delta_l\}_{l=1}^{\mathsf{L}}$ are not all strictly positive or negative, then \texttt{DYCHEM} can generate coherent forecasts.}

\textbf{Remark} ~Since the probability that all $\{\delta_l\}_{l=1}^{\mathsf{L}}$ have same sign decrease exponentially as we increase $\mathsf{L}$, we say that the more diverse forecasting models we employed in \texttt{DYCHEM}, the more likely the assumption will hold. In other words, the increased number of heterogeneous models brings more robustness to the forecasts. This is in contrast to the single model case in SHARQ, where a stronger unbiased assumption at certain level is required to obtain coherent results. Comparing with MinT which strictly enforces forecasts to be coherent, \texttt{DYCHEM} provides a controllable trade-off between accuracy on each time series and coherency over the given hierarchy, leading to better results in empirical evaluations.

% Note that other reconciliation methods like MinT, requires forecasting model at every vertex to be unbiased. This is not a realistic assumption given the likelihood of changing dynamics at different aggregation levels. SHARQ also requires unbiasedness assumption at the bottom level vertex. However, the assumption of \texttt{DYCHEM} is weaker as it only requires the true value to lie in the convex hull formed by forecasts from each expert. Therefore, the increased number of heterogeneous experts brings more robustness to the forecasts.

\subsection{Model-Free Quantile Estimations}
Adding uncertainty estimations to point forecasts provide a more comprehensive view of prediction. However, combining a diverse set of forecasting models could negatively impact the probabilistic forecasts, as the combined distribution will become more ``spread out'' or underconfident \citep{ranjan2010combining}. Directly estimating quantiles is an ideal solution, but not each model (particularly for encapsulated APIs) can perform quantile regression freely given its working mechanism. A viable approach is to build a quantile estimation module that is independent to the forecasting models, while solely dependent on the point prediction results. In other words, we perform quantile estimation on the ``black-box'' point prediction.

% We resort to quantiles for characterizing uncertainty. Since there is neither distributional nor model assumptions in our framework, one cannot simply compute the quantiles of a given distribution, or combine the quantile loss from the base models. The target of uncertainty measurement is therefore a black-box. Specifically, for each vertex $v_i$, one only has access to the input $x^{v_i}_{1: t_i}$ and output $\hat{x}^{v_i}_{t_i+1: T_i}$ of the MoE framework. 

We first introduce conditional quantile regression for our forecasting problem, which is used to train the quantile estimators. For each quantile value $\tau \in [0, 1]$ and input $x_{1: t_0}$, the aim of quantile regression is to estimate the $\tau^{th}$ quantile function $q(\tau, x_{1: t_0}) := \inf \{\hat{x}_t\in \R: F(\hat{x}_t~|~x_{1: t_0}, \Theta) \geq \tau\},$ where $F$ is the CDF function of $P(\hat{x}_t~|~x_{1: t_0}, \Theta)$ for any forecasting time stamp $t \in [t_0, T]$. The quantile estimation at each $\tau$ is achieved by minimizing the pinball loss function 
% $\L_q$ defined by
\begin{equation}
    \L_q(x_{1: T}, \tau) = \sum_{t=t_0}^T (x_t - q(\tau, x_{1: t_0})) \cdot (\tau - \mathbbm{1}[x_t < q(\tau, x_{1: t_0})]), \nonumber
    % \label{eq:qloss}
\end{equation}
where $\mathbbm{1}[\cdot]$ is the indicator function. Ideally, the estimated quantiles $q(\tau, x_{1: t_0})$ are monotonically increasing w.r.t. $\tau$. We call it quantile crossing if this condition is not satisfied. Quantile crossing is a common problem when $n \geq 2$ quantile estimators are evaluated by minimizing e.g., $\sum_{s=1}^n \L_q(x_{1: T}, \tau_s)$. One way to address this is to represent quantile estimator as $$q(\tau_s, x_{1:t_0}) = \int_0^{\tau_s} \phi(y, x_{1: t_0}; \Theta_q)~dy + q(0, x_{1:t_0}) ~~ \forall \tau_s \in (0, 1],$$ where $\phi(y, x_{1: t_0}; \Theta_q)$, the derivative of $q(\tau_s, x_{1:t_0})$, is parameterized by neural networks ($\Theta_q$) whose outputs are made to be strictly positive. This is to make sure the quantile estimator is monotonically increasing on $\tau_s$. However, the integral dependency on $\tau_s$ limits only one quantile can be estimated at a time. We now introduce a new form of quantile estimator, where a straightforward improvement is it only requires one-time computation of integral to generate all quantiles. Specifically, we can estimate the set of quantiles $\{q(\tau_s, x_{1: t_0})\}_{s=1}^n, ~ \forall \tau_s \in [0, 1]$ by first computing
% The core idea is to model the derivative of quantile estimations using neural networks:
% \begin{equation}
%     \{q(\tau_s, x^{v_i}_{1: t_i}, \hat{x}^{v_i}_{t_i+1: T_i}; \Theta_u)\}_{s=1}^n, ~ \forall \tau_s \in [0, 1] \leftarrow \int_{-1}^{1} \phi(t, x^{v_i}_{1: t_i}; \Theta_u) ~dt + C_0(x^{v_i}_{1: t_i}, \hat{x}^{v_i}_{t_i+1: T_i}),
%     \label{eq:uncertainty}
% \end{equation}
\begin{equation}
    \Phi = \int_{-1}^{1} \phi(y, x_{1: t_0}; \Theta_q) ~dy + C_0,
    \label{eq:uncertainty}
\end{equation}
where $\phi$ is still the function approximator that is set to be strictly positive, and $[-1, 1]$ is the range of $2\tau_s - 1$ with $\tau_s \in [0, 1]$. We can then obtain each $q(\tau_s)$ through $f(\Phi, \tau_s)$, where $f$ is a pre-defined function. However, we need to bypass the integral operation in Eq (\ref{eq:uncertainty}) since it cannot be trained by the quantile loss. We then replace the integral expression of $q(\tau_s)$ using its Chebyshev approximation \citep{clenshaw1955note}: similar to Taylor approximation for polynomials, Chebyshev approximation is a commonly used basis in numerical integration. By approximating $q(\tau_s)$, we obtain the following equation 
\begin{equation}
    q(\tau_s) = \sum_{k=1}^{d-1} C_k T_k(2\tau_s - 1) + C_0 \quad \tau_s \in [0, 1].
\end{equation}
$T_k: [-1, 1] \mapsto \R$ is the Chebyshev polynomial, $\{C_k\}_{k=0}^{d - 1}$ are the corresponding Chebyshev coefficients, and $d$ is the degree of approximation, where higher $d$ results in better approximation. Since $\{T_k\}_{k=0}^{d-1}$ are recurrently defined and don't need to be explicitly computed, one only needs to calculate the Chebyshev coefficients to obtain $q(\tau_s)$. Specifically, $\{C_k\}_{k=1}^{d-1}$ can be obtained by applying a linear transformation on neural network $\phi$ followed by some recurrent computations. After $\{C_k\}_{k=1}^{d-1}$ is determined, we compute $C_0$ using the point forecasting results $\hat{x}_{t_0:T}$, which connects $\hat{x}_{t_0:T}$ to the distribution formed by the set of quantiles $\{q(\tau_s)\}_{s=1}^n$.

\begin{figure*}[t!]
    \begin{minipage}{0.97\textwidth}
    \centering
    \begin{tabular}{@{\hspace{-2.4ex}} c @{\hspace{-1.5ex}} @{\hspace{-2.4ex}} c @{\hspace{-1.5ex}} @{\hspace{-2.4ex}} c @{\hspace{-1.5ex}} @{\hspace{-2.4ex}} c @{\hspace{-2.4ex}}}
        \begin{tabular}{c}
        \includegraphics[width=.27\textwidth]{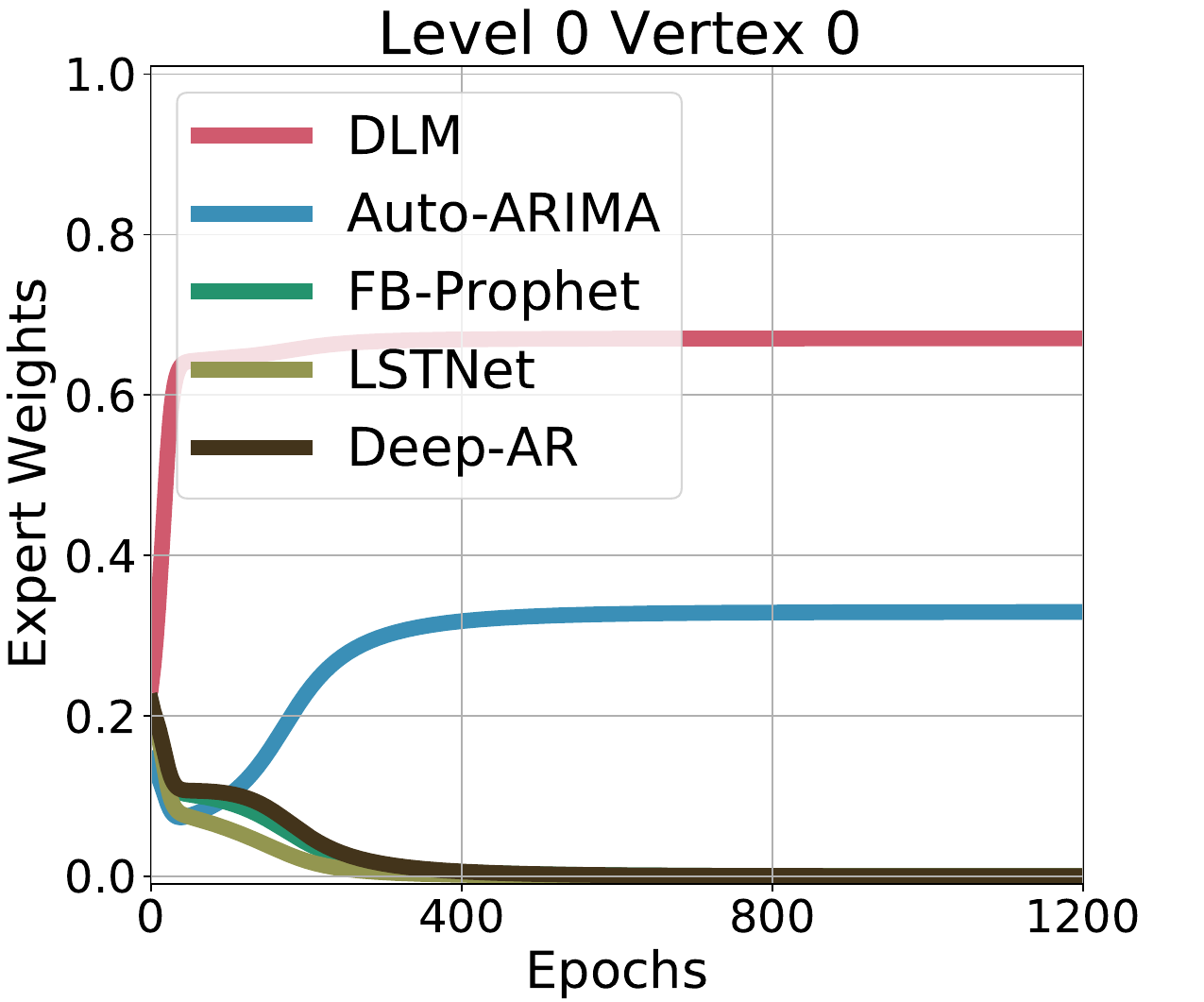}
        %\vspace{-5pt}
        \\
        {\small{(a)}}
        \end{tabular} &
        \begin{tabular}{c}
        \includegraphics[width=.27\textwidth]{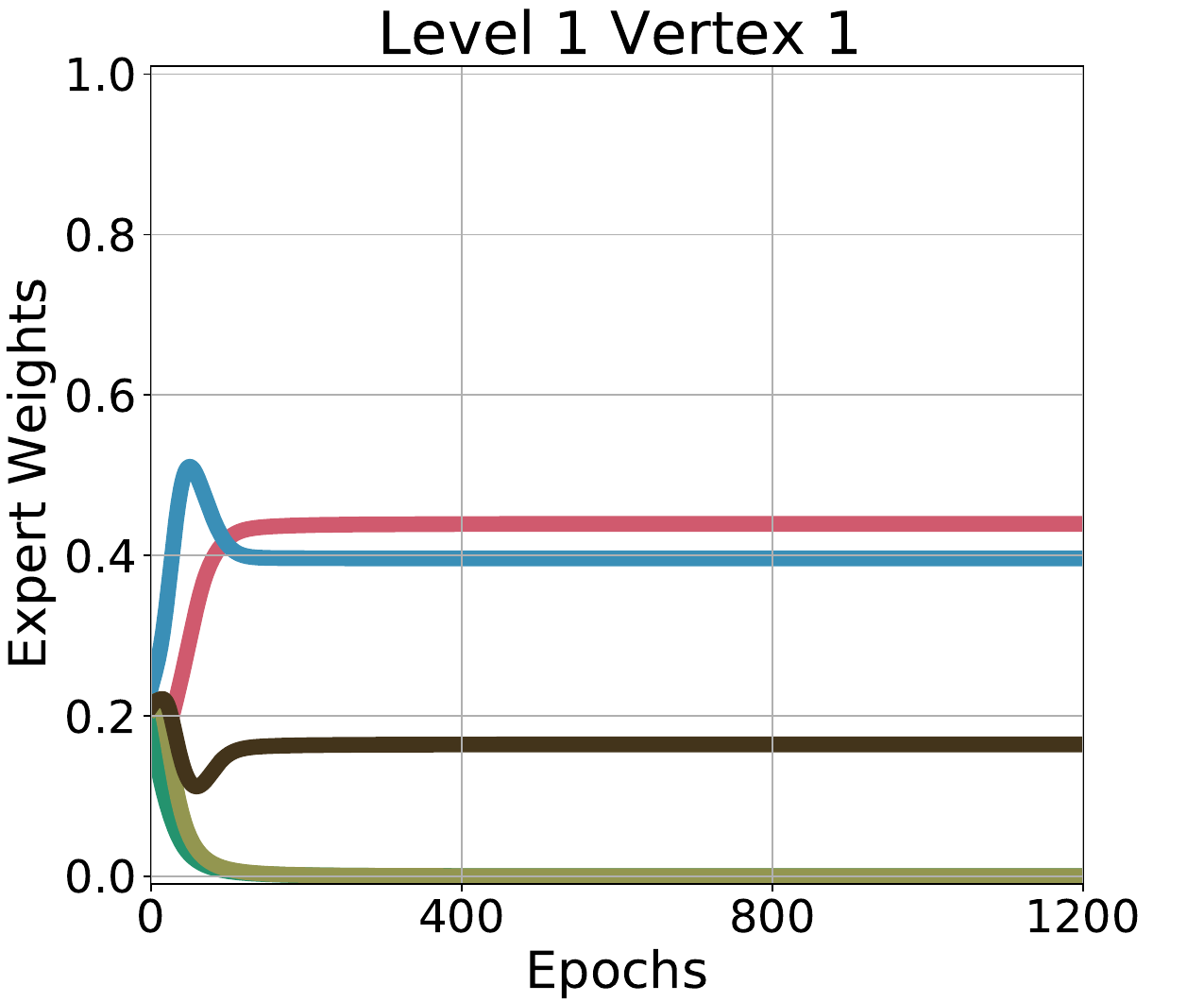}
        %\vspace{-5pt}
        \\
        {\small{(b)}}
        \end{tabular} & 
        \begin{tabular}{c}
        \includegraphics[width=.27\textwidth]{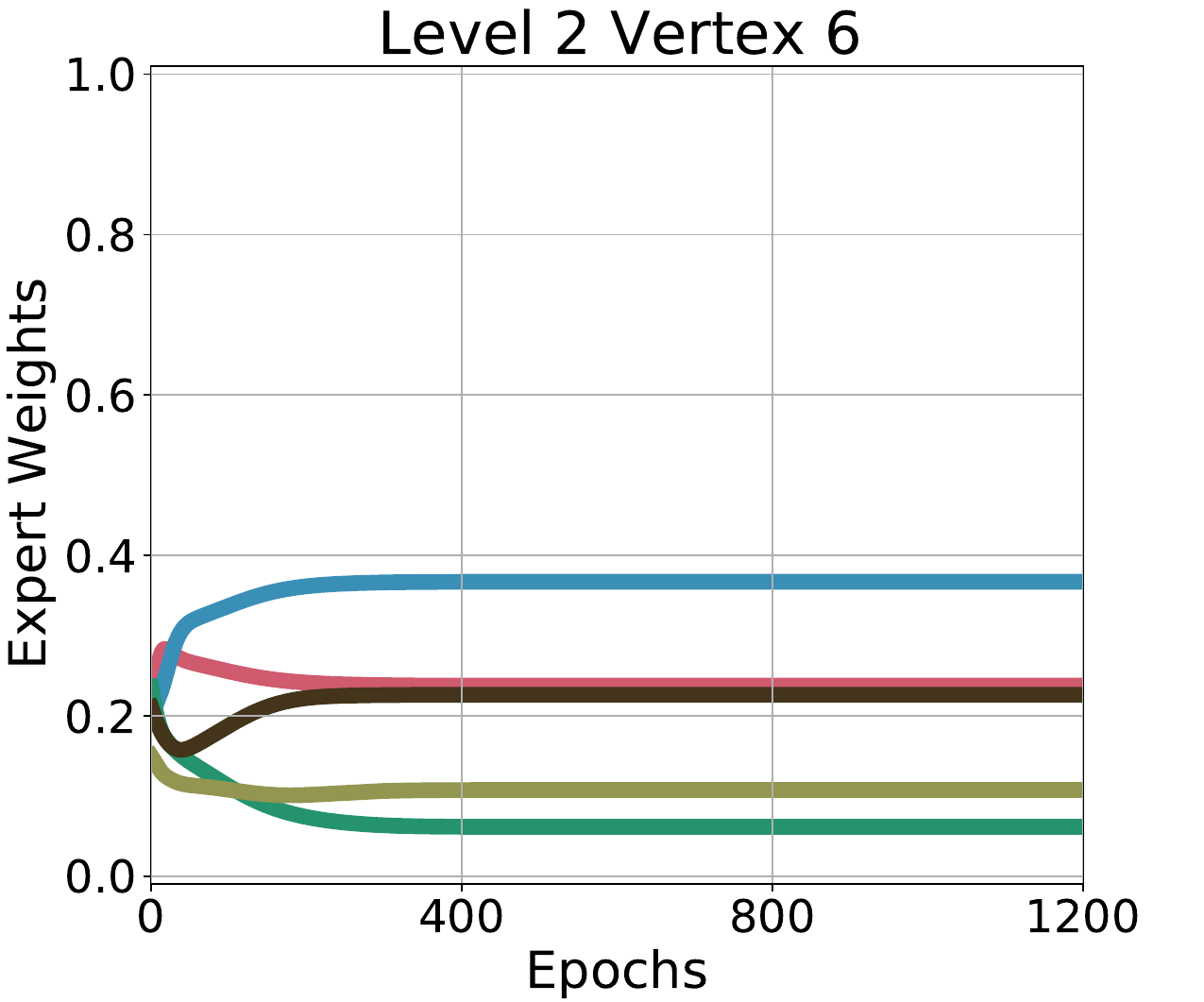} 
        %\vspace{-5pt}
        \\
        {\small{(c)}}
        \end{tabular} &
        \begin{tabular}{c}
        \includegraphics[width=.27\textwidth]{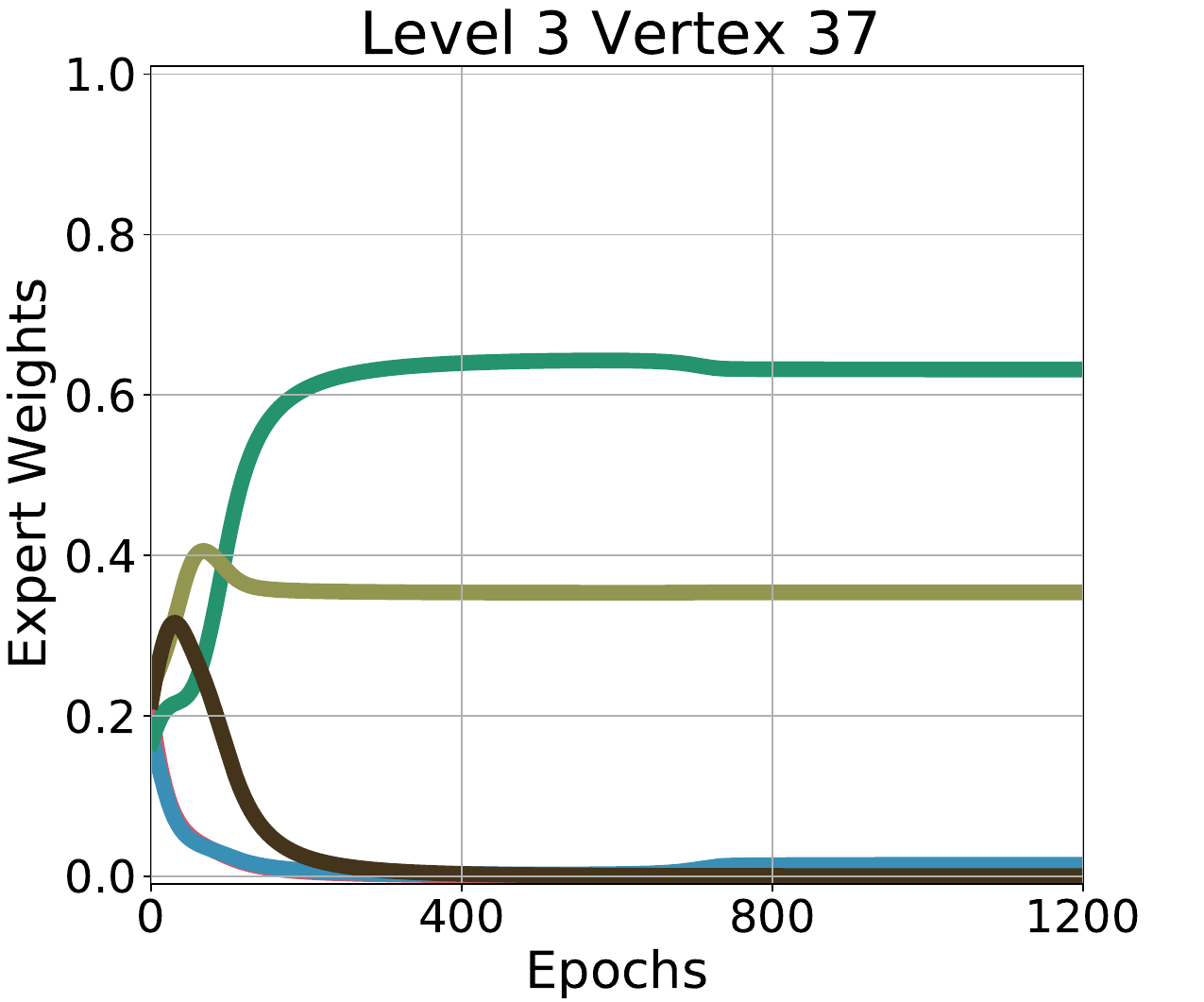}
        %\vspace{-5pt}
        \\
        {\small{(d)}}
        \end{tabular} \\
        \end{tabular}
    \end{minipage}
    \caption{Pre-trained base models' weight curve from vertices at different aggregation levels. The most suitable representation of each time series can be found in diverse combinations of base models.}
    \label{fig:gating_weight}
\end{figure*}

\textbf{Proposition 2} [\textit{Connection to Point Forecast}] \label{prop:point}
\textit{By setting the Chebyshev coefficient $C_0$ as
\begin{equation}
    C_0 = 2\cdot \hat{x}_{t_0: T} - 2\sum_{k=1, k~\mathrm{even}}^{d-1} (-1)^{k/2} ~C_k,
    \label{eq:median}
\end{equation}
the point forecast $\hat{x}_{t_0:T}$ is the median of the distribution formed by the set of quantiles $\{q(\tau_s)\}_{s=1}^n$.
}

\textbf{Remark} ~Connecting to $\hat{x}_{t_0:T}$ calibrates quantile estimations to be consistent with point forecasts, which have been regularized to be near coherent across the hierarchy.
% In fact, one can impose the point forecast $\hat{x}_{t_0:T}$ to be any summary statistic (e.g., mean, max, min) of the distribution and the equation of $C_0$ can be derived accordingly. 

% One can also combine the approximator with Eq (\ref{eq:qloss}) to be trained using gradient-based methods. Moreover, the important advantages are the model-free property and its capability of generating multiple quantile estimations solely based on the point forecasting results. See figure 2 right...
% transform the estimation of quantile functions into estimation of Chebyshev coefficients, and this is followed by a general transformation procedure, which waive the model dependency...

In summary, \texttt{DYCHEM}'s quantile estimation framework is to transform the direct estimation of $\{q(\tau_s)\}_{s=1}^n$ into estimation of its components, Chebyshev coefficients, one of which ($C_0$) can be used to bind with point forecasts. We only require the input $x_{1:t_0}$ (for training function approximator $\phi$) and output $\hat{x}_{t_0:T}$ (for computing coefficient $C_0$) of the point prediction module and are therefore not dependent on which point prediction model we choose. Compared with directly estimating $\{q(\tau_s)\}_{s=1}^n$, this procedure improves quantile estimations at each level by
1. eliminating quantile crossing; 2. improving quantile coherency: they are both difficult in regular quantile regression. Figure \ref{fig:gating} wraps up the overall structure of \texttt{DYCHEM}. 

%% file: tex/experiments.tex
\section{Experiments} \label{sec:exp}

In this section, we provide a comprehensive evaluation of \texttt{DYCHEM}. Our experiments include: 1. coherent forecasts on hierarchically aggregated time series data (section \ref{sec:point} - \ref{sec:coherency}); 2. real-time forecasting under change of time series dynamics (section \ref{sec:cp}); 3. forecasting massive financial records on an industry forecasting pipeline \ref{sec:financial}. Our results show that \texttt{DYCHEM} can significantly improve forecasting performance over baselines and is adaptive to change of dynamics. Implementations of \texttt{DYCHEM} along its baselines, datasets and hyper-parameters of all following experiments can be found at \href{https://github.com/aaronhan223/htsf}{\textcolor{magenta}{github.com/aaronhan223/htsf}}.

\begin{table*}[t!]
\centering
\renewcommand\arraystretch{0.8}
\scalebox{1}{
\begin{tabular}{c|c|c|c|c|c|c}
\hlinewd{1.5pt}
Data $\backslash$ Method & Level & DYCHEM-LOO & SHARQ & HIER-E2E & Average & DYCHEM \\ \hline
\multirow{4}{*}{ Labour } & 1 & 43.33 $\pm$0.42 (\texttt{.054}) & 52.07 $\pm$0.45 (\texttt{.085}) & 45.12 $\pm$0.23 (\texttt{.085}) & 49.34 $\pm$0.65 (\texttt{.075}) & \textbf{38.84 $\pm$0.04} (\textbf{\texttt{.045}}) \\
& 2 & 53.68 $\pm$0.68 (\texttt{.104}) & 58.69 $\pm$0.41 (\texttt{.120}) & 55.61 $\pm$0.74 (\texttt{.107}) & 60.87 $\pm$0.33 (\texttt{.119}) & \textbf{48.64 $\pm$0.78} (\textbf{\texttt{.092}}) \\
& 3 & 57.16 $\pm$0.25 (\texttt{.135}) & 64.02 $\pm$0.09 (\textbf{\texttt{.132}}) & 60.03 $\pm$0.26 (\texttt{.134}) & 69.29 $\pm$0.42 (\texttt{.138}) & \textbf{49.17 $\pm$0.36} (\texttt{.144}) \\
& 4 & 65.05 $\pm$0.18 (\textbf{\texttt{.153}}) & 72.13 $\pm$0.34 (\texttt{.167}) & 71.38 $\pm$0.15 (\texttt{.154}) & 75.56 $\pm$0.94 (\texttt{.156}) & \textbf{61.22 $\pm$0.14} (\texttt{.163}) \\ \hline
\multirow{4}{*}{ M5 } & 1 & 49.29 $\pm$0.34 (\texttt{.071}) & 56.31 $\pm$0.17 (\texttt{.054}) & 51.69 $\pm$0.05 (\texttt{.070}) & 59.61 $\pm$0.38 (\texttt{.104}) & \textbf{42.61 $\pm$0.14} (\textbf{\texttt{.046}}) \\
& 2 & 54.36 $\pm$0.28 (\texttt{.127}) & 62.16 $\pm$0.27 (\textbf{\texttt{.079}}) & 54.72 $\pm$0.63 (\texttt{.116}) & 60.48 $\pm$0.58 (\texttt{.133}) & \textbf{49.75 $\pm$0.22} (\texttt{.084}) \\
& 3 & 55.18 $\pm$0.22 (\texttt{.142}) & 65.37 $\pm$0.63 (\texttt{.134}) & 65.02 $\pm$0.24 (\texttt{.142}) & 68.29 $\pm$0.25 (\texttt{.143}) & \textbf{53.61 $\pm$0.42} (\textbf{\texttt{.101}}) \\
& 4 & 59.04 $\pm$0.36 (\texttt{.164}) & 72.86 $\pm$0.27 (\texttt{.189}) & 72.04 $\pm$0.36 (\texttt{.164}) & 70.29 $\pm$0.34 (\texttt{.168}) & \textbf{57.89 $\pm$0.47} (\textbf{\texttt{.109}}) \\ \hline
\multirow{4}{*}{ AEDemand } & 1 & 61.35 $\pm$0.76 (\texttt{.132}) & 68.19 $\pm$0.29 (\textbf{\texttt{.113}}) & 64.45 $\pm$0.48 (\texttt{.213}) & 67.32 $\pm$0.29 (\texttt{.164}) & \textbf{59.89 $\pm$0.32} (\texttt{.145}) \\
& 2 & 58.12 $\pm$0.46 (\texttt{.152}) & 66.57 $\pm$0.24 (\texttt{.199}) & 63.72 $\pm$0.36(\texttt{.131}) & 63.58 $\pm$0.72 (\texttt{.129}) & \textbf{55.72 $\pm$0.73} (\textbf{\texttt{.122}}) \\
& 3 & 66.38 $\pm$0.78 (\texttt{.124}) & 68.25 $\pm$0.47 (\texttt{.131}) & 68.01 $\pm$0.22 (\texttt{.126}) & 70.44 $\pm$0.09 (\texttt{.124}) & \textbf{62.55 $\pm$0.14} (\textbf{\texttt{.111}}) \\
& 4 & 76.58 $\pm$0.63 (\texttt{.136}) & 87.35 $\pm$0.69 (\texttt{.225}) & 82.47 $\pm$0.28 (\texttt{.192}) & 73.22 $\pm$0.37 (\texttt{.135}) & \textbf{71.45 $\pm$0.43} (\textbf{\texttt{.125}}) \\ \hline
\multirow{5}{*}{ Wiki } & 1 & 65.98 $\pm$0.22 (\texttt{.121}) & 70.36 $\pm$0.24 (\texttt{.147}) & 69.67 $\pm$0.58 (\textbf{\texttt{.067}}) & 66.42 $\pm$0.16 (\texttt{.128}) & \textbf{63.27 $\pm$0.73} (\texttt{.117}) \\
& 2 & 68.54 $\pm$0.47 (\texttt{.157}) & 73.06 $\pm$0.42 (\texttt{.159}) & 68.24 $\pm$0.33 (\textbf{\texttt{.108}}) & 72.01 $\pm$0.52 (\texttt{.157}) & \textbf{65.14 $\pm$0.46} (\texttt{.143}) \\
& 3 & 72.42 $\pm$0.36 (\texttt{.149}) & 76.15 $\pm$0.34 (\textbf{\texttt{.135}}) & 74.62 $\pm$0.19 (\texttt{.155}) & 74.37 $\pm$0.83 (\texttt{.147}) & \textbf{69.48 $\pm$0.33} (\texttt{.156}) \\
& 4 & 77.12 $\pm$0.23 (\texttt{.268}) & 78.42 $\pm$0.34 (\texttt{.201}) & 79.63 $\pm$0.41 (\texttt{.291}) & 81.38 $\pm$0.65 (\texttt{.278}) & \textbf{75.69 $\pm$0.76} (\textbf{\texttt{.189}}) \\
& 5 & 84.77 $\pm$0.49 (\texttt{.241}) & 85.12 $\pm$0.62 (\texttt{.345}) & 79.65 $\pm$0.24 (\texttt{.326}) & 84.68 $\pm$0.42 (\texttt{.221}) & \textbf{76.88 $\pm$0.72} (\textbf{\texttt{.213}}) \\ \hlinewd{1.5pt}
\end{tabular}}
\caption{Forecasting performance measured by averaged $\mathrm{MASE}^{\downarrow}$ and $\mathrm{CRPS}^{\downarrow}$ (within bracket) on 4 HTS datasets. M5 has more than 4 levels, we report results on the top 4 levels due to space constraint. All experiments are repeated 5 times.}
\label{tab:mape}
\end{table*}

% \vspace{-2ex}
\begin{figure*}[t!]
    \begin{minipage}{\textwidth}
    \centering
    % \vspace{-2ex}
    \begin{tabular}{@{\hspace{-2.4ex}} c @{\hspace{-1.5ex}} @{\hspace{-2.4ex}} c @{\hspace{-1.5ex}} @{\hspace{-2.4ex}} c @{\hspace{-1.5ex}}}
        \begin{tabular}{c}
        \includegraphics[width=.35\textwidth]{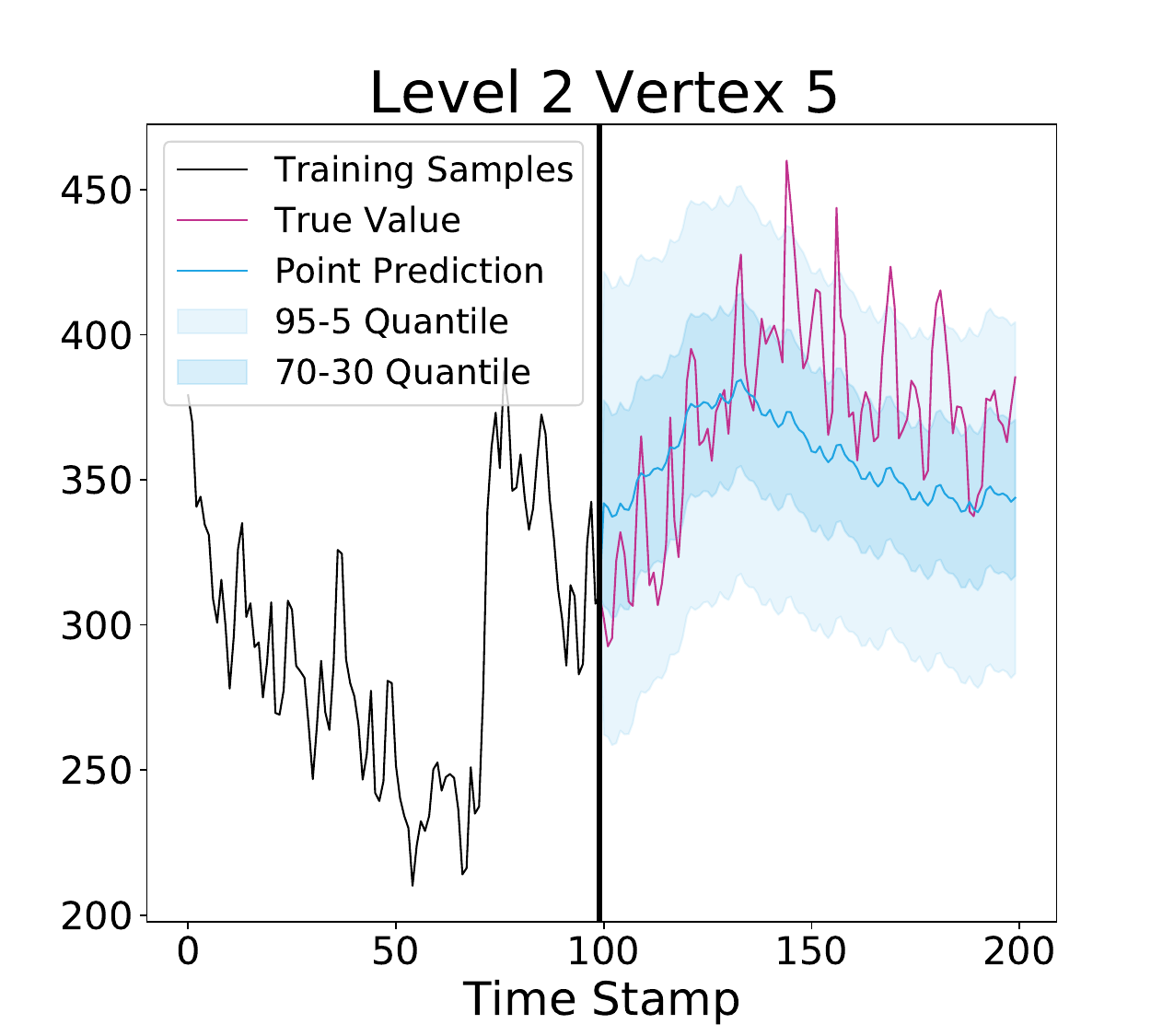}
        %\vspace{-5pt}
        \\
        {\small{(a)}}
        \end{tabular} &
        \begin{tabular}{c}
        \includegraphics[width=.35\textwidth]{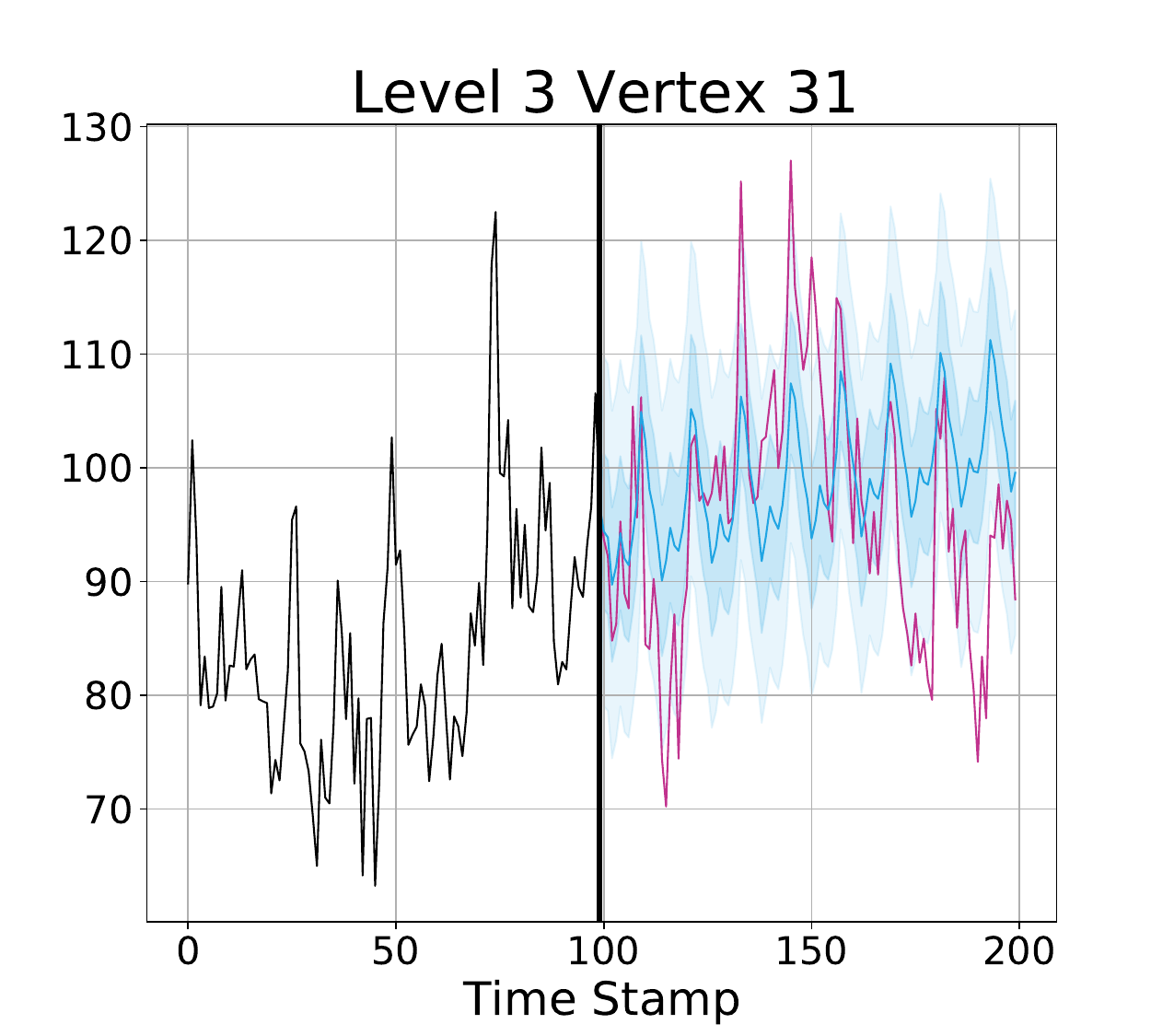}
        %\vspace{-5pt}
        \\
        {\small{(b)}}
        \end{tabular} & 
        \begin{tabular}{c}
        \includegraphics[width=.35\textwidth]{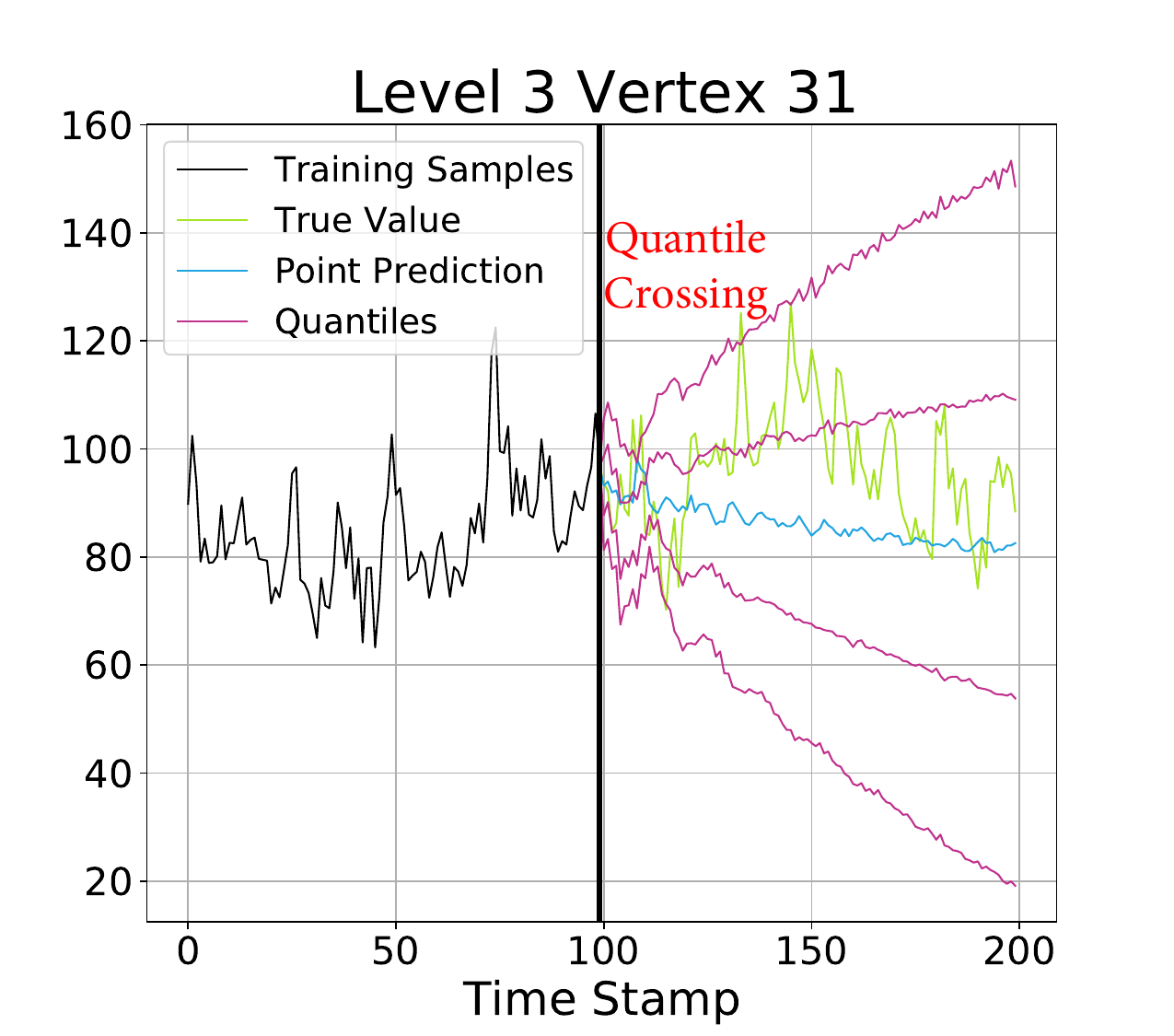} 
        %\vspace{-5pt}
        \\
        {\small{(c)}}
        \end{tabular} \\
        \end{tabular}
    \end{minipage}
    \vspace{-2ex}
    \caption{(a), (b) Quantile forecasting results generated by \texttt{DYCHEM} at vertex 5 and 31 of the Australian Labour data, where $\tau_s = [0.05, 0.3, 0.5, 0.7, 0.95]$. (c) SHARQ at same $\tau_s$, results showing mild quantile crossing.}
    \label{fig:forecasts}
\end{figure*}

\textbf{Evaluation Metrics}
~We employ a commonly used relative error measure called mean absolute scaled error (MASE) \citep{hyndman2006another} to evaluate point forecast accuracy. We use continuous ranked probability score (CRPS) \citep{matheson1976scoring} to measure quantile estimations, where it has been shown that the CRPS and quantile loss are equivalent \citep{si2021autoregressive}. In addition, we define a coherency loss to measure the coherency of time series forecast on a given hierarchy with $m$ vertices
\begin{equation}
    \mathrm{Coherency~ Loss} = \frac{1}{h}\sum_{t=T}^{T+h}\sum_{i=1}^m |\hat{x}_t^{v_i} - \sum_{e_{i, k} \in E} \hat{x}_t^{v_k}|.
    \label{eq:cl}
\end{equation}

\textbf{Choosing Diverse Base Models}
~Models for time series span a wide range of categories. 
\texttt{ARIMA(p,d,q)} \citep{hyndman2008automatic} is able to model many time series with trend and seasonality components, but its performance is normally poor on long-term forecast and time series with change-points. Dynamic regression models \citep{taylor2018forecasting} extend ARIMA by inclusion of other external variables, but the process requires expertise and experimentation. If all variables are concurrent, difficulties in forecasting external variables will result in poor forecasts.
% State-space models \citep{durbin2012time, hyndman2008forecasting, west1985dynamic} provides a more general and interpretable framework for modeling time series by sequentially updating information to give better estimates; Kalman filter \citep{welch1995introduction} and exponential smoothing \citep{hyndman2008forecasting} are both prominent examples. 
Deep Neural Networks \citep{lai2018modeling, salinas2020deepar} can improve the ability to model complex data with enough history. However, it is difficult to obtain a single model that works well in all situations. We choose 5 commonly used forecasting models that span across the above-mentioned categories as the base model of \texttt{DYCHEM}, including dynamic linear models (DLM) \citep{west2006bayesian}, Auto-ARIMA \citep{hyndman2008automatic}, Facebook Prophet \citep{taylor2018forecasting}, LSTNet \citep{lai2018modeling} and DeepAR \citep{salinas2020deepar}. Although there is no strict rule on how to choose base models, it is beneficial to diversify forecasting models as they can be complementary in most situations. Since LSTNet and DeepAR are global models \citep{januschowski2020criteria}, which learn across a set of time series, we only need to train one model for a certain hierarchy. The rest of the models are trained on univariate time series, where we assign one model for each vertex. In the following experiments, we will show that the cooperation of these models will lead to significant improvement in forecasting.

% It is defined by
% \begin{equation}
%     \mathrm{MASE} = \frac{\frac{100}{h}\times \sum_{t=T}^{T+h}|x_t - \hat{x}_t|}{\frac{1}{T-1}\sum_{t=2}^T|x_t - x_{t-1}|}.
%     \label{eq:mase}
% \end{equation}

% Specifically, CRPS measures the compatibility of a cumulative distribution function $F$ with an observation $x$ as:
% \begin{equation}
%     \mathrm{CRPS}(F, x) = \int_{\mathbb{R}} (F(z) - \mathbb{I}\{x \leq z\})^2 ~dz,
%     \label{eq:crps}
% \end{equation}
% where $\mathbb{I}\{x \leq z\}$ is the indicator function. CRPS attains its minimum when the distribution $F$ and the distribution formulated by $x$ (can be either scalar or vector) are equal. All CRPS results reported in the experiment are one-step ahead forecasting results.

% \paragraph{Public Datasets}

\subsection{Point Forecast} \label{sec:point}
\textbf{Experiment Setting}
% We choose 5 commonly used forecasting models as the base model of \texttt{DYCHEM}, they are dynamic linear models (DLM) \citep{west2006bayesian}, Auto-ARIMA \citep{hyndman2008automatic}, Facebook Prophet \citep{taylor2018forecasting}, LSTNet \citep{lai2018modeling} and DeepAR \citep{salinas2020deepar}. Since LSTNet and DeepAR are global models \citep{januschowski2020criteria}, which learn across a set of time series, we only need to train one model for a certain hierarchy. The rest of the models are trained on univariate time series, where we assign one model for each vertex. In other words, one can include both global models and univariate models in \texttt{DYCHEM}. However, we will not discuss how to make the best selection of each heterogeneous model, which is beyond the scope of this paper. 
~We compared with SHARQ \citep{han2021simultaneously} (LSTM as the base model), HIER-E2E \citep{rangapuram2021end}, and ensemble averaging \citep{bhatnagar2021merlion}. In addition, to show that a diverse set of base models brings more robustness to forecast, we perform leave-one-out (LOO) for every base model of \texttt{DYCHEM} and results are averaged across each run. We conducted our experiments on 4 public time-series datasets with hierarchical aggregations: Australian Labour\footnote[1]{https://www.abs.gov.au/statistics}, M5 \citep{makridakis2020m5}, Wikipedia\footnote[2]{https://www.kaggle.com/c/web-traffic-time-series-forecasting} and AEDemand \citep{athanasopoulos2017forecasting}. 

% \textcolor{red}{no universal way of deciding the best set of forecasting models but the guiding principle is falling into diverse set of categories...}

% \textcolor{red}{individual contribution of each component...}

Figure \ref{fig:gating_weight} illustrates the weight evolution for each model as the gating network is trained and reconciled on vertices at each level using the Australian Labour dataset. According to the plots, the gating network emphasizes DLM for time series at the top aggregated level; as the level becomes lower, DLM becomes less dominant and other models begin to carry more weights at some vertices. Finally, at a bottom-level vertex where data becomes noisy, different models take charge of the forecast. The plots demonstrates individual contribution of each expert and the best way to represent different aggregated levels of HTS are distinct. Intuitively, learning weights from a richer space can improve the results over simple averaging. Table \ref{tab:mape} shows level-wise point forecasting results measured by MASE. \texttt{DYCHEM} generates better results over baselines that includes single forecasting model (SHARQ and HIER-E2E), LOO for every base model, as well as combining models by ensemble averaging. The results shows that combining heterogeneous models through a gating network can significantly improve the forecasting accuracy. From Table \ref{tab:mape}, we can also see that time series at lower aggregated levels are harder to forecast, given it is more likely to have sparsity issue.

% Specifically, SHARQ \citep{han2021simultaneously} and \texttt{MECATS} are single model and ME combined with in-training reconciliation, respectively. For point forecast, ME can also be combined with post-training reconciliation methods by removing the regularization term in Eq (\ref{eq:loss}) and adding reconciliation steps after training. 
% Figure \ref{fig:gating_weight} the gating network is trained on Australian Labour dataset 

% (\textcolor{red}{replot}).
% \vspace{-1ex}
\subsection{Quantile Forecasts} \label{sec:quantile}
% \vspace{-0.5ex}
We now evaluate the quantile forecasts after point forecasts are obtained. The neural network models $\phi$ are fed into mini-batches of training data $\mathrm{sw}(x_{1:t_0})$ generated by sliding windows, which are used to approximate Chebyshev coefficients. Each quantile estimation can then be computed recurrently and trained by quantile loss function.
% Implementation details and sudo code for training each quantile can be found at Appendix \ref{sec:app_che}. 
Figure \ref{fig:forecasts} demonstrates quantile forecasting results on some arbitrary chosen vertices in Austrlian Labour dataset. Overall, the multi-quantile forecasts can well-capture the distribution of future time stamps at different aggregation levels, while the point forecasts binds the other quantiles by serving as the median of distribution. As a comparison, we also show an example of SHARQ trained with LSTM model, where the multiple quantile estimations are less constrained by the point forecast. Note that SHARQ is a very strong baseline, providing results superior to specific observation noise model such as Gaussian or negative binomial distribution. However, quantile crossing is possible since SHARQ cannot guarantee strict monotonicity w.r.t $\tau_s$. Table \ref{tab:mape} quantitatively compares each method using CRPS: although the quantile forecasting results didn't reach the same performance as point forecasts, \texttt{DYCHEM} still outperforms other baselines in the majority of situations.
% the proposed method is combined with ME (the last row), while other single models are trained with SHARQ. Note that MinT approaches are originally applied on point forecasts, our uncertainty wrappers are then built on top of the post-processed point estimations. Overall, the proposed multi-quantile generator captures the uncertainties in a better way.

% Since $\phi$ needs to be evaluated at different $t_k$, 
% Since post-reconciliation methods can only work with point prediction. Long-term, multi-step prediction... CRPS \citep{matheson1976scoring}
% \vspace{-2ex}
\begin{figure*}[t!]
    \begin{minipage}{\textwidth}
    \centering
    \begin{tabular}{@{\hspace{-2.4ex}} c @{\hspace{-1.5ex}} @{\hspace{-2.4ex}} c @{\hspace{-1.5ex}} @{\hspace{-2.4ex}} c @{\hspace{-1.5ex}}}
        \begin{tabular}{c}
        \includegraphics[width=.35\textwidth]{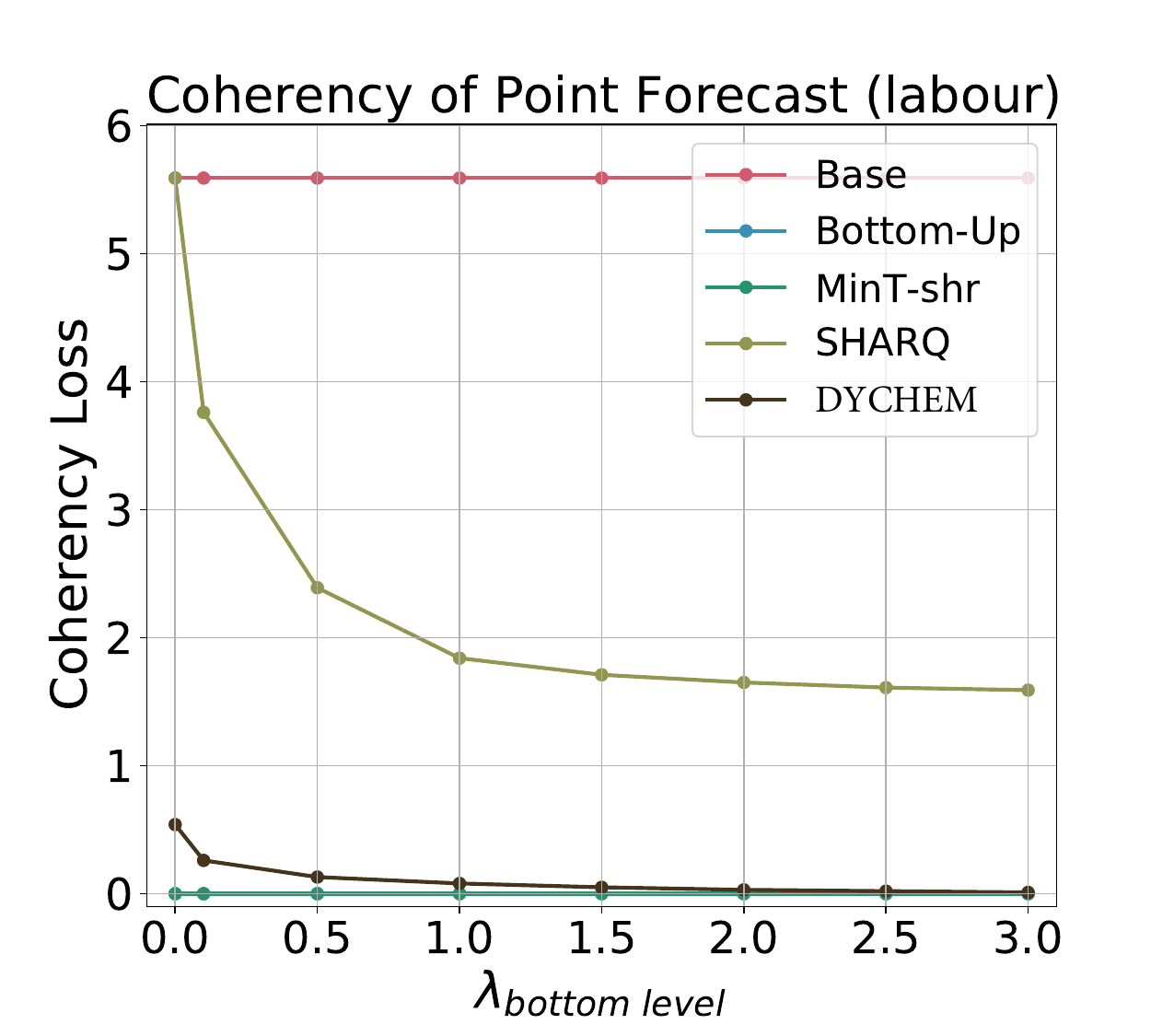}
        %\vspace{-5pt}
        \\
        {\small{(a)}}
        \end{tabular} &
        \begin{tabular}{c}
        \includegraphics[width=.35\textwidth]{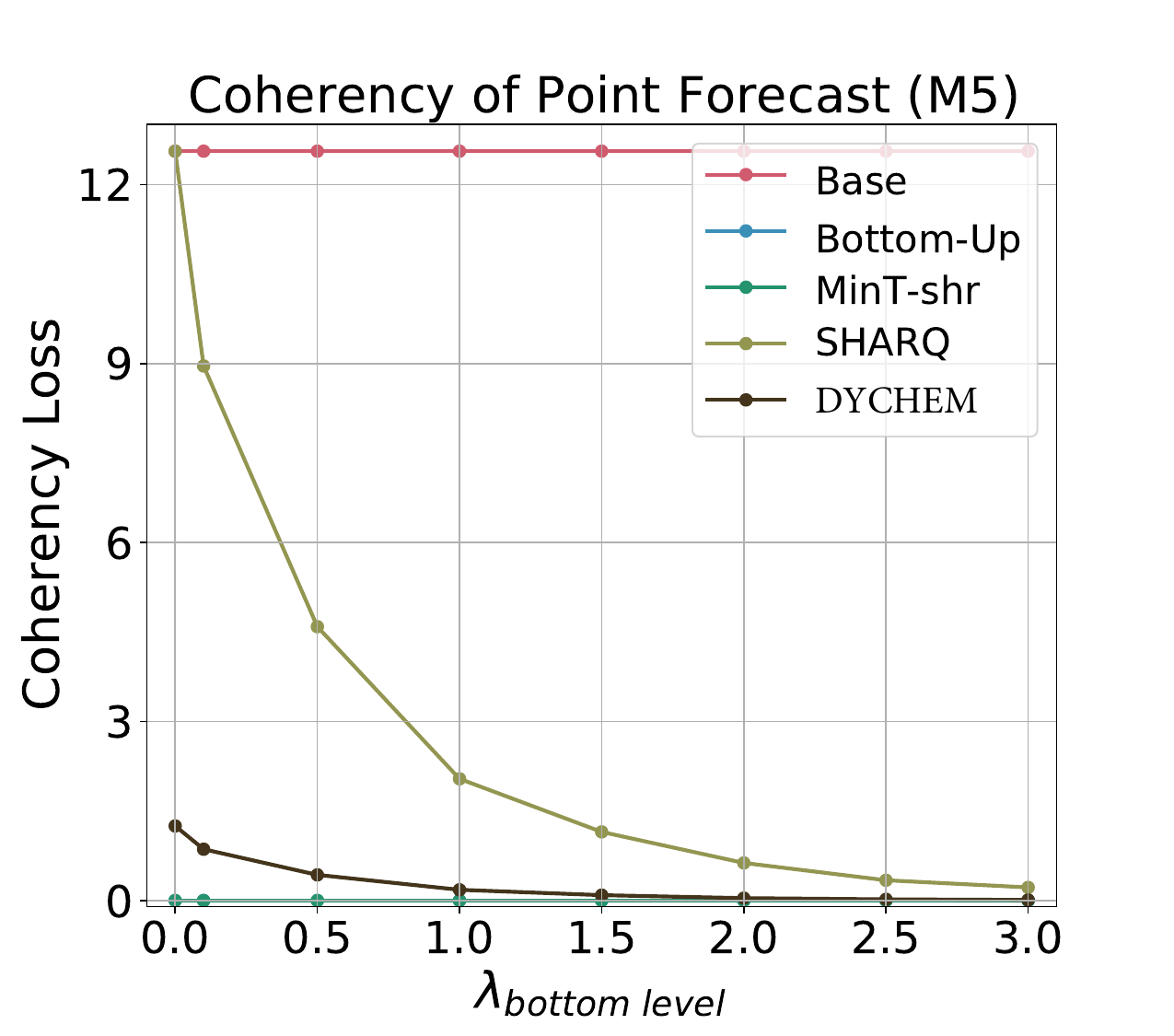}
        %\vspace{-5pt}
        \\
        {\small{(b)}}
        \end{tabular} & 
        \begin{tabular}{c}
        \includegraphics[width=.38\textwidth]{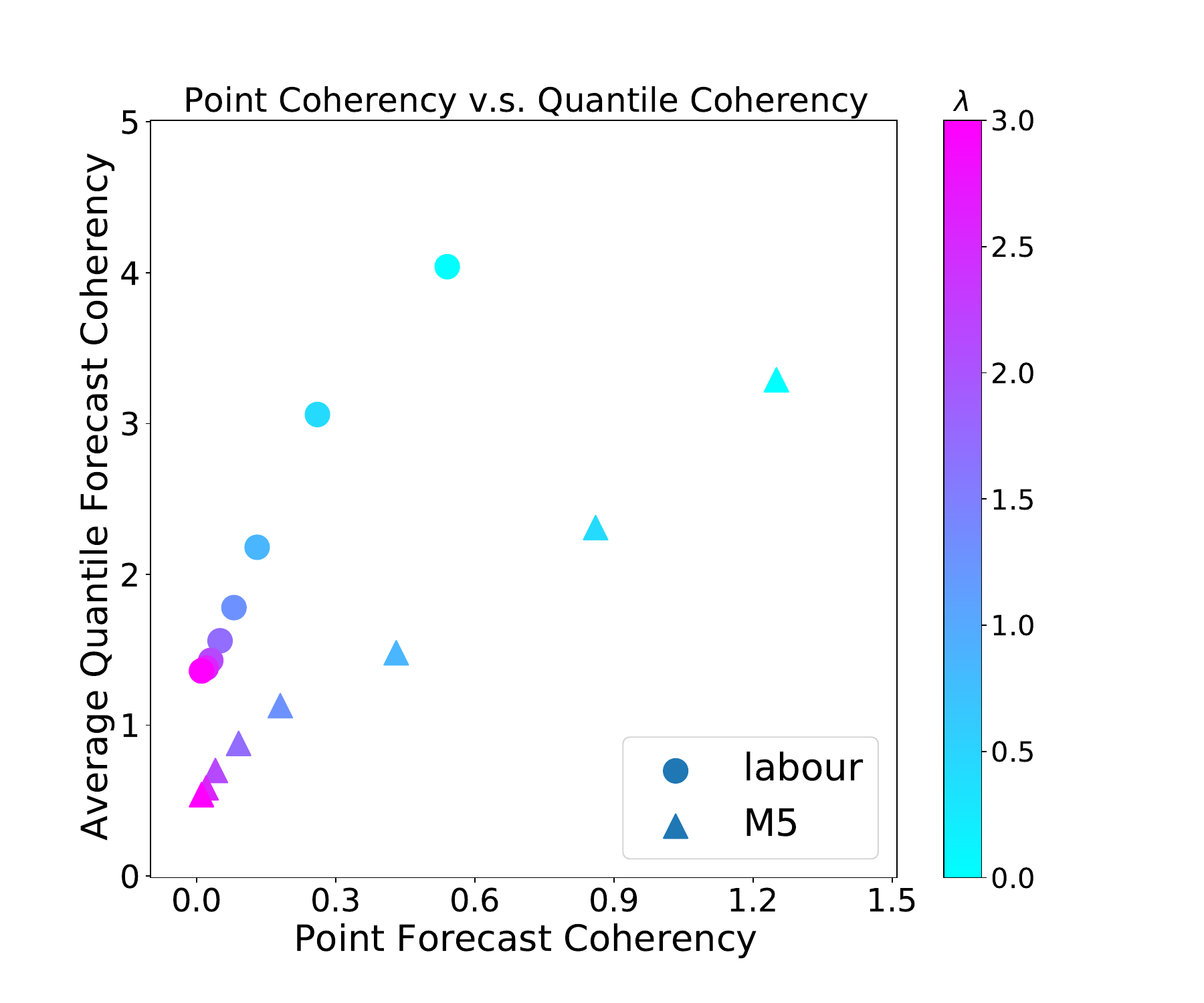} 
        %\vspace{-5pt}
        \\
        {\small{(c)}}
        \end{tabular} \\
        \end{tabular}
    \end{minipage}
    \vspace{-2ex}
    \caption{(a), (b) Absolute coherency loss of point forecast w.r.t. regularization strength $\lambda$ on Australian Labour and M5 dataset. (c) Relationship between point forecast coherency and average of quantile coherency.}
    \label{fig:coherency}
\end{figure*}

\subsection{Coherency Analysis} \label{sec:coherency}
In our empirical studies, it is common that the absolute coherency loss defined by (\ref{eq:cl}) cannot be completely eliminated, due to various reasons such as inappropriate choice of models or bad training of gating networks. We tune parameter $\lambda$ in Eq (\ref{eq:loss}) that controls the penalty strength for coherency using bottom-level time series. For HTS with more than two aggregation levels, we progressively reduce $\lambda$ at higher levels based on the value of $\lambda_{\mathrm{bottom~level}}$, as higher level time series are easier to calibrate. We observe that by choosing $\lambda_{\mathrm{bottom~level}}$ in an appropriate range, coherency loss can be mitigated without sacrificing forecasting accuracy. We also compare with a few other baselines including Base method: regular time series forecast without reconciliation; Bottom-Up method: only forecast bottom-level time series and aggregate according to hierarchical structure to obtain higher level forecasts; MinT-shr: MinT with shrinkage estinator for covariance matrix. Figure \ref{fig:coherency} (a), (b) have shown that \texttt{DYCHEM} can achieve nearly coherent results. In fact, while \texttt{DYCHEM} significantly improves the accuracy of point forecast, it can also positively affect the coherency. Note that post-processing methods such as MinT, can reduce the coherency loss to a very small value at the cost of decreased accuracy, since the summation of bottom-level forecasts is ``forced'' to be aligned with higher-level forecasts. Figure \ref{fig:coherency} (c) shows that the improved coherency of point forecast can positively affect the coherency of quantiles. This result corresponds to our claim on Proposition 2. However, since the additive property does not hold for quantiles \citep{han2021simultaneously}, there still exists non-zero coherency loss in certain quantiles.

\begin{figure}[t]
    \centering
    \vspace{-1ex}
    \includegraphics[width=.75\linewidth]{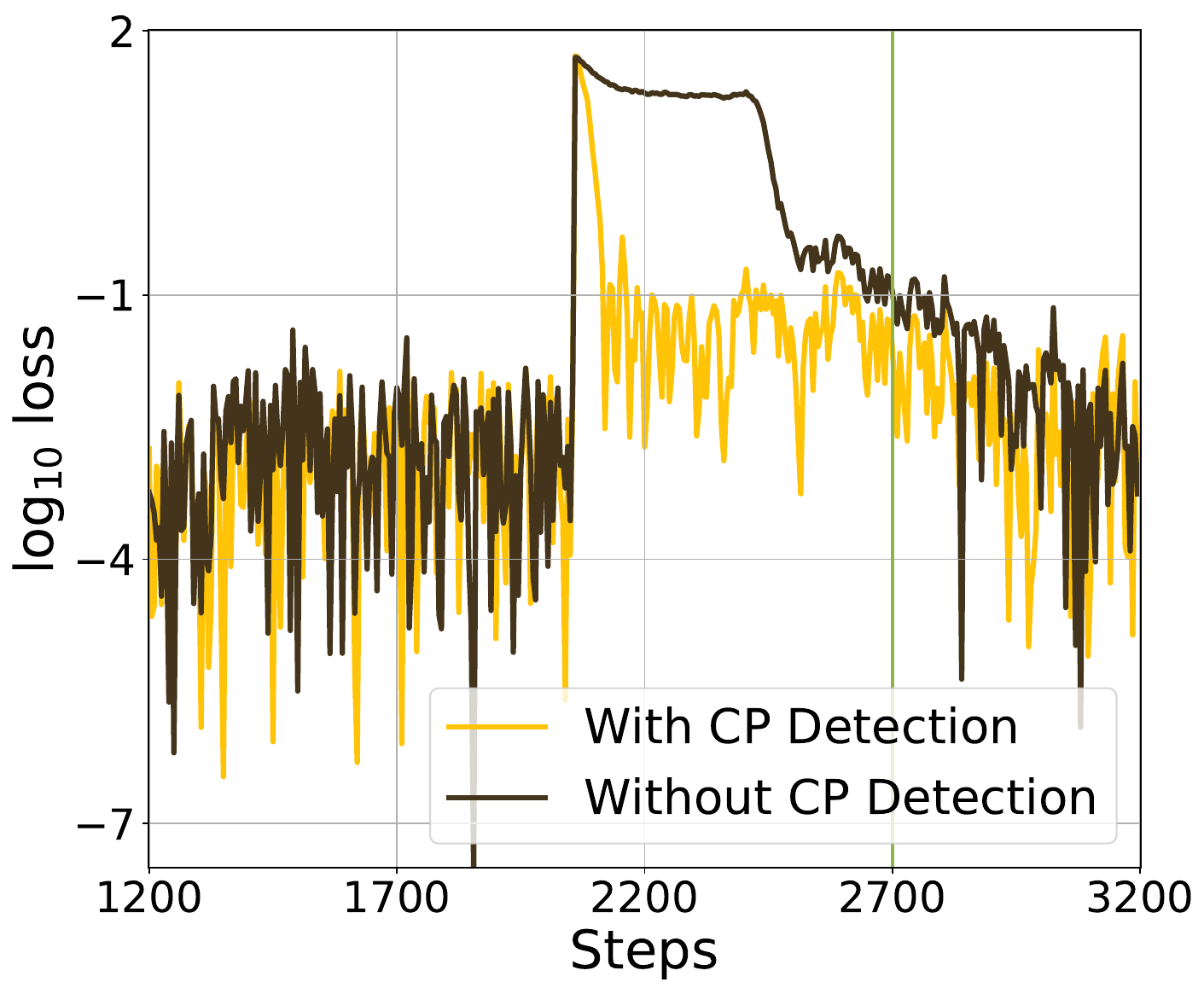}
    \caption{Online time series forecasting using \texttt{DYCHEM} under change-points.}
    % \vspace{-2ex}
    \label{fig:sim_cp}
\end{figure}

\subsection{Forecasting Under Change of Dynamics} \label{sec:cp}
In many applications, samples are collected sequentially, which requires the model to be updated in an \textit{online} manner. It is possible that the change of dynamics will occur in such settings. We show that \texttt{DYCHEM} is robust when this is happened. Since gating network has been proven to be reasonably adaptive to abrupt changes \citep{chaer1997mixture}, we further alleviate the jump of loss during the transient period in adapting to new dynamics with the help of Bayesian online change-point detection (BOCPD) \citep{adams2007bayesian}: after detecting change-points, we combine the original weights $\{w_l\}_{l=1}^{\mathsf{L}}$ with averaged weights $1/ \mathsf{L}$ using a shrinkage factor that decreases exponentially through time. We assume new samples come with a batch size of 1; both gating network and models' parameters are updated at each step. Fig. \ref{fig:sim_cp} shows the behaviour of \texttt{DYCHEM} when change-point occurs at around step 2050 and finally adapts to the new dynamics at step 2700.
% Table \ref{tab:sim_cp} (a) demonstrates the original gating networks' behavior when change-point occurs at step 2050 (step 1200 is the starting point of online updates) on simulated data. The network adapts to the new dynamics at around step 2700. 
With the help of BOCPD, MSE loss during this transient period can be greatly reduced. 
% We follow the procedure in the change-point detection literature \citep{adams2007bayesian, knoblauch2018spatio} to evaluate our method on several real-world sequential data with change-points, where table \ref{tab:sim_cp} (c) shows that our method can detect most changes. In addition, by combining \texttt{DYCHEM} with BOCPD (\texttt{DYCHEM-CP}), the predictive results to the next step are from combining multiple forecasting models instead of a single forecasting model in BOCPD. Table \ref{tab:sim_cp} (d) shows that the averaged predictive loss of \texttt{DYCHEM-CP} also outperforms a similar work BOCPD-MS \citep{knoblauch2018spatio} that adds model selection to BOCPD.

\subsection{Forecasting Massive User Financial Records} \label{sec:financial}
\texttt{DYCHEM} has been tested within a large financial software company for cash flow forecasting. The hierarchy consists of 5 vertices, where the top level vertex represents the total expense of one user, the four bottom vertices are the first top, second top, third top and the rest of expenses of the corresponding user. We conducted experiments on 12,000 users' data that possess the same hierarchy, and compared with already deployed baseline methods (PyDLM\footnote[1]{https://github.com/wwrechard/PyDLM}, PyDLM + MinT-shr) using normalised root mean squared error (NRMSE). \texttt{DYCHEM} significantly outperforms the deployed baselines. In addition, we have also modified \texttt{DYCHEM} for industrial deployment, which parallelized the training of forecasting models and gating networks at the same aggregated level given their mutual independence. Implementation and results can be found via the link provided in the beginning of section \ref{sec:exp}.

%% file: tex/conclusion.tex
\section{Discussion} \label{sec:limit}

Forecasting hierarchically aggregated time series is an understudied problem with many applications. In this work, we propose \texttt{DYCHEM} that learns to combine a set of heterogeneous forecasting models while also performs quantile estimations in a model-free manner. \texttt{DYCHEM} has demonstrated superior results in both accuracy and coherency in HTS forecasting, and is adaptive to change of dynamics in sequential data. We also observed improvement of performance when deployed \texttt{DYCHEM} into an industry forecasting pipeline. In summary, \texttt{DYCHEM} produces forecasts that are more reliable, which are critical requirements for widespread adoption and safe deployment. The substantially more flexibility that \texttt{DYCHEM} offers in terms of using a customized mix of heterogeneous models for improving a specific time series forecast comes at an increased computational cost. It also introduces a few more hyper-parameters. Our future research will focus on ameliorating the costs when a large number of forecasts need to be done. In particular, we would like to further investigate situations where data resources are more constrained, such as short sequences or sequences with missing entries. 

% \paragraph{Ethics Statement}
% Forecasts that are coherent and accurate, adaptive to changes in the nature of the time series, and also yield
% confidence intervals, result in solutions that are more reliable and trustworthy, critical requirements for widespread adoption and safe deployment. So our social impact is mostly positive since robustness and reliability of a predictive model are important aspects of responsible AI solutions. In addition, using a modular structure like mixture of heterogeneous experts may allow the use of simpler base models that are more interpretable by humans, without sacrificing the accuracy of the overall solution. In addition, this work does not involve human subjects, and also does not raise any discrimination/bias/fairness concerns. All information of public datasets, models and hyper-parameters can be found in Appendix \ref{sec:app_f}. 

% \paragraph{Reproducibility}
% We have submitted the source code of this work in a zip file as part of the supplementary materials. Implementation instructions can also be found under the same file.

%% file: tex/appendix.tex
\appendix
\definecolor{grn}{RGB}{27, 129, 62}
% \section{Proof of Proposition 1} \label{sec:app_a}
% \textbf{Proposition 1.} [\textit{Coherency of \texttt{DYCHEM}}]
% \textit{Given $\mathsf{L}$ models are assigned to $v_1$ and its child vertices $\{v_i\}_{i=2}^n$. The models generate point forecast $\{\hat{x}_{v_i}^l\}_{l=1}^{\mathsf{L}}$ at $v_i$, where $\hat{x}_{v_i}^l \in \R$ is not necessarily unbiased. For all $\mathsf{L}$ models, assume their coherency loss $\{\delta_l\}_{l=1}^{\mathsf{L}}$ are not all strictly positive or negative, then \texttt{DYCHEM} can generate coherent forecasts.}
% \vspace{2ex}
\textbf{A. Proof of Proposition 1}
\begin{proof}
For each forecasting model, we obtain the following hierarchical forecasts which are not necessarily unbiased
\begin{equation}
	\hat{x}_{v_1}^1 = \sum_{i=2}^n \hat{x}_{v_i}^1 + \delta_1; \quad \hat{x}_{v_1}^2 = \sum_{i=2}^n \hat{x}_{v_i}^2 + \delta_2 ~~ \dots ~~ \hat{x}_{v_1}^{\mathsf{L}} = \sum_{i=2}^n \hat{x}_{v_i}^{\mathsf{L}} + \delta_{\mathsf{L}}.
	\label{eq:coherency}
\end{equation}
With out loss of generality, we assume $\hat{x}_{v_i}^1 \leq \hat{x}_{v_i}^2 \leq \dots \leq \hat{x}_{v_i}^{\mathsf{L}}$. The gating network generates a set of weights $\{w^l_{v_i}\}_{l=1}^{\mathsf{L}} \in [0, 1]$ for models at vertex $v_i$ and $\{w_l\}_{l=1}^{\mathsf{L}} \in [0, 1]$ on each coherency loss $\delta_l$ correspondingly. We can then rewrite Eq (\ref{eq:coherency}) as follows
\begin{equation}
	\sum_{l=1}^{\mathsf{L}} w_{v_1}^l\hat{x}_{v_1}^l = \sum_{i=2}^n\sum_{l=1}^{\mathsf{L}} w_{v_i}^l\hat{x}_{v_i}^l + \sum_{l=1}^{\mathsf{L}} w_l \delta_l.
	\label{eq:comb}
\end{equation}
Since $\{\delta_l\}_{l=1}^{\mathsf{L}}$ cannot be all strictly positive or negative, we can therefore find a set of weights $\{w_l\}_{l=1}^{\mathsf{L}} \in [0, 1]$ such that $\sum_{l=1}^{\mathsf{L}} w_l \delta_l = 0,$ so \texttt{DYCHEM} is coherent.
\end{proof}

\textbf{Remark} An equivalent statement of $\{\delta_l\}_{l=1}^{\mathsf{L}}$ do not have same sign is the forecasting ground truth value $x_{v_i} \in [\hat{x}_{v_i}^1, \hat{x}_{v_i}^{\mathsf{L}}], \forall i \in [1, n]$. In other words, if $x_{v_i}$ can be written as the convex combination of each model's forecast, \texttt{DYCHEM} can generate coherent forecasts. The convex hull assumption is much weaker than unbiased assumption used by other methods, making \texttt{DYCHEM} more robust to different applications.

% The ideal value of $w_l$ in Eq (\ref{eq:comb}) can be written as $w_l = \frac{w_{v_1}^l\hat{x}_{v_1}^l - \sum_{i=2}^n w_{v_i}^l\hat{x}_{v_i}^l}{\hat{x}_{v_1}^l - \sum_{i=2}^n \hat{x}_{v_i}^l}.$ By minimizing $\mathcal{L}_{\mathrm{recon}}$, the gating network is trained to generate weights $\{w_{v_i}^l\}_{l=1}^{\mathsf{L}}$ such that $\sum_{l=1}^{\mathsf{L}} w_{v_i}^l \hat{x}_{v_i}^l = x_{v_i},$ since $x_{v_i} \in [\hat{x}_{v_i}^1, \hat{x}_{v_i}^{\mathsf{L}}]$ can be written as the convex combination of each model's forecast. The generated weights should also satisfy $w_{v_1}^l\hat{x}_{v_1}^l - \sum_{i=2}^n w_{v_i}^l\hat{x}_{v_i}^l = w_l \delta_l$. There is normally a trade-off between accuracy on each time series and coherency over the given hierarchy in empirical evaluations. 
\vspace{2ex}
\textbf{B. Chebyshev Approximation} \label{sec:app_che}
\paragraph{General Procedure}
In order to train neural network $\phi$ using gradient-based methods, we first need to approximate the integral $\int_{-1}^{1} \phi(y) ~dy$. We can transform this problem by a change of variable: $\int_{0}^{\pi} \phi(\cos \theta) \sin \theta ~d\theta$. Define the Chebyshev polynomial $T_k: [-1, 1] \rightarrow \R$ and its recurrent relationship 
% $T_0(z) := 1, ~T_1(z) := z, ~T_{k+1}(z) := 2zT_k(z) - T_{k-1}(z), ~ k \geq 1.$
\begin{equation}
    T_0(z) = 1, ~ T_1(z) = z, ~ T_{k+1}(z) = 2zT_k(z) - T_{k-1}(z), ~ k \geq 1,
    \label{eq:recurrent}
\end{equation}
which can also be written as $T_k(\cos \theta) = \cos(k\theta)$. By definition, we can write $\phi(\cos \theta)$ as its Chebyshev polynomial approximation \citep{clenshaw1955note}: $\phi(\cos \theta) = \frac{1}{2} c_0 + \sum_{k=1}^{\infty} c_k \cdot T_k(\cos \theta)$, where $\{c_k\}_{k=0}^{\infty}$ are coefficients. Normally, a finite number of terms can achieve sufficient precision for approximation, e.g.
\begin{equation}
    \phi(\cos \theta) \approx \frac{1}{2} ~c_0 + \sum_{k=1}^{d-1}~ c_k \cdot T_k(\cos \theta)
    \label{eq:approx}
\end{equation}
is a $d$ degree Chebyshev polynomial approxiamtion of $\phi(\cos \theta)$. Therefore we have
\begin{align*}
\int_{0}^{\pi} \phi(\cos \theta) \sin \theta ~d\theta &= \int_{0}^{\pi} \phi(\cos \theta) ~d\cos \theta \\
&= \int_{-1}^1 \left[ \frac{1}{2} ~c_0 + \sum_{k=1}^{d-1}~ c_k \cdot T_k(z) \right]~dz, \nonumber
\end{align*}
where $z = 2\tau_s - 1, \forall \tau_s \in [0, 1]$. By the recurrent definition in Eq (\ref{eq:recurrent}), the integral of the Chebyshev polynomial $T_k$ corresponds to a new Chebyshev polynomial:
$
\int T_0(z)~dz = T_1(z), ~ \int T_1(z)~dz = \frac{T_2(z)}{4} - \frac{T_0(z)}{4}, ~ \int T_k(z)~dz = \frac{T_{k-1}(z)}{2(k-1)} - \frac{T_{k+1}(z)}{2(k+1)}.
$
Therefore, the integration result can also be written in terms of Chebyshev polynomials: $\Phi(z) = \frac{1}{2}C_0 + \sum_{k=1}^{d-1}C_k\cdot T_k(z)$, where the new coefficients $\{C_k\}_{k=1}^{d-1}$ can be obtained from the original coefficients by the following equations
\begin{equation}
    C_k = \frac{c_{k-1} - c_{k+1}}{4k}, ~ 0<k<d-1, \quad
    C_{d-1} = \frac{c_{d-2}}{4(d-1)}.
    \label{eq:coeff}
\end{equation}
The above derivation shows that the integral operation in Eq (\ref{eq:uncertainty}) can be replaced by Chebyshev polynomial approximation:
\begin{equation}
    q(\tau_s) = \sum_{k=1}^{d-1}C_k\cdot T_k(2\tau_s -1) + C_0.
    \label{eq:approx_2}
\end{equation}

Since the polynomials $T_k$ can be obtained in a recurrent manner instead of explicitly computed \citep{clenshaw1955note}, estimating Chebyshev coefficients $\{c_k\}_{k=0}^{d-1}$ in Eq (\ref{eq:approx}) is an important step.

\begin{algorithm}[t]
\small
\caption{Compute Chebyshev Coefficients}
\label{alg:cheby}
\begin{algorithmic}
\STATE \textbf{Input}: time series data $\mathrm{sw}(x_{1:t_0})$, Chebyshev polynomial degree $d$, MLP $\phi(\Theta_q)$, point forecast $\hat{x}_{t_0: T}$, batch size $\mathrm{bs}$, window length $\omega$
\STATE \textbf{Process}:
\STATE $\{y_k\}_{k=0}^{d-1} = \cos \left(\frac{\pi (k+\frac{1}{2})}{d}\right), ~~ k \in [0, d)$ ~\textcolor{grn}{$\rhd$ Chebyshev roots}
\STATE $\mathcal{T}_k \leftarrow \texttt{Repeat}(\{y_k\}_{k=0}^{d-1}, \lceil \frac{\mathrm{bs}}{d} \rceil + 1)$ ~\textcolor{grn}{$\rhd$ repeat $y_k$ vector $\lceil \frac{\mathrm{bs}}{d} \rceil + 1$ times}
\STATE $\mathrm{sw}(\mathcal{T}_k) \in \R^{bs \times d}$ ~\textcolor{grn}{$\rhd$ generate $d$ vectors of roots by rolling window on $\mathcal{T}_k$ with step size 1}
\FOR{$j=0, \dots, d-1$}
\STATE $X_j \leftarrow \texttt{concat}[\mathrm{sw}(\mathcal{T}_k)[:, j], ~~\mathrm{sw}(x_{1:t_0})] \in \R^{\mathrm{bs} \times (\omega + 1)}$ ~\textcolor{grn}{$\rhd$ combine different root values with time series and feed into model}
\STATE $O_j \leftarrow \phi_j(X_j, \Theta_q)$
\STATE $P_j \leftarrow \texttt{softplus}(O_j + 1e^{-5}) + 1e^{-3}$ ~\textcolor{grn}{$\rhd$ strictly positive outputs}
\ENDFOR
\STATE $\{c_k\}_{k=0}^{d-1} \leftarrow \texttt{DCT}(\{P_j\}_{j=0}^{d-1})$ 
\STATE $\{C_k\}_{k=0}^{d-1} \leftarrow C_k = \frac{c_{k-1} - c_{k+1}}{4k}, ~~ 0<k<d-1, \quad C_{d-1} = \frac{c_{d-2}}{4(d-1)}$
\STATE $C_0 = 2\cdot \hat{x}_{t_0: T} - 2 \sum_{k=1, k ~\mathrm{even}}^{d-1} (-1)^{k/2} C_k$
\STATE \textbf{Return}: $\{C_k\}_{k=0}^{d-1}$
\end{algorithmic}
\end{algorithm}

\paragraph{Compute Chebyshev Coefficients}
% The coefficients can be obtained by a linear transformation of the set of neural network outputs $\{\phi(y_k, x_{1: t_0}; \Theta_q)\}_{k=0}^{d-1}$ evaluated at $\{y_k\}_{k=0}^{d-1}$, where $y_k = \cos \left(\frac{\pi (k+\frac{1}{2})}{d}\right)$, $k \in [0, d)$ are the Chebyshev roots. The linear transformation ($\R^d \rightarrow \R^d$) is implemented by the Discrete Cosine Transform (DCT) algorithm, which gives $\{c_k\}_{k=0}^{d-1}$ by transforming the set of output scalars.

We evaluate $\phi(y, x_{1: t_0}; \Theta_q)$ at its $d$ uniformly distributed roots for $d$-dimension truncated Chebyshev polynomial approximation, i.e., $\{\phi(y_k, x_{1: t_0}; \Theta_q)\}_{k=0}^{d-1}$. During implementation, we first use the sliding window approach to process the input data $x_{1: t_0}$, where $\mathrm{sw}(x_{1:t_0})$ can be processed by $\phi(\Theta_q)$ in a common supervised learning way. $\{y_k\}_{k=0}^{d-1}$ serves as an additional feature of the input of $\phi(\Theta_q)$, where $\phi(\Theta_q): \R^{\mathrm{bs}\times (\omega+1)} \rightarrow \R^{\mathrm{bs}\times 1}$ is implemented as a multilayer perceptron (MLP) with strictly positive output, where $\mathrm{bs}$ and $\omega$ represent the batch size and window length, respectively. Since the MLP should be evaluated at multiple roots, we combine $\mathrm{sw}(x_{1:t_0}) \in \R^{\mathrm{bs}\times \omega}$ with different $\{t_k\}_{k=0}^{d-1}$ and feed them into $d$ replicates of MLP respectively. The outputs of $d$ MLP are then transformed by the Discrete Cosine Transform (DCT) algorithm to obtain the coefficients $\{c_k(x_{1: t_0})\}_{k=0}^{d-1}$. Full procedure in computing the Chebyshev coefficients can be found in Algorithm \ref{alg:cheby}, which returns the set of coefficients $\{C_k\}_{k=0}^{d-1}$ with size $\R^{\mathrm{bs}\times d}$. These coefficients are then used to compute quantile values by iteratively combining with $T_k$ defined in Eq (\ref{eq:recurrent}), where $z = 2\tau_s - 1 \in [-1, 1]$ and $\tau_s$ can be any specified quantiles. 

% In summary, \texttt{DYCHEM} shows some very good properties compared to previous methods. It can borrow strength from arbitrary forecasting models in a plug-and-play manner, leading to accurate and robust forecasts in all situations. The connection between its point forecasts and quantile estimations can also improve the performance of probabilistic forecasts. Additionally, \texttt{DYCHEM} tackles model dependency and quantile crossing problems that have not been addressed by SHARQ and also does not need strong assumptions made by the two-stage reconciliation methods.

\vspace{2ex}
\textbf{C. Proof of Proposition 2}
% \textbf{Proposition 2} [\textit{Connection to Point Forecast}] 
% \textit{By setting the Chebyshev coefficient $C_0$ as
% \begin{equation}
%     C_0 = 2\cdot \hat{x}_{t_0: T} - 2\sum_{k=1, k~\mathrm{even}}^{d-1} (-1)^{k/2} ~C_k,
% \end{equation}
% the point forecast $\hat{x}_{t_0:T}$ is the median of the distribution formed by the set of quantiles $\{q(\tau_s)\}_{s=1}^n$.
% }
\begin{proof}
If $\hat{x}_{t_0: T}$ is the median estimation, we have $q(0.5, x_{1: t_0}) = \hat{x}_{t_0: T}$, since $T_k(0) = 0$ when $k$ is odd and $T_k(0) = (-1)^{k/2}$ when $k$ is even, and 
\begin{equation}
    q(0.5, x_{1: t_0}) = \sum_{k=1}^{d-1}C_k(x_{1: t_0})\cdot T_k(0) + \frac{1}{2} C_0(x_{1: t_0}, \hat{x}_{t_0: T}),
\end{equation}
we can therefore obtain the value of $C_0$ given the point forecasts are median:
\begin{equation}
    C_0(x_{1: t_0}, \hat{x}_{t_0: T}) = 2\cdot \hat{x}_{t_0: T} - 2\sum_{k=1, k~\mathrm{even}}^{d-1} (-1)^{k/2} C_k(x_{1: t_0}).
    \label{eq:median_app}
\end{equation}
\end{proof}

\textbf{Remark} 
Similarly, if $\hat{x}_{t_0: T}$ is determined to be the mean prediction, by definition we have
\begin{align}
    \int_{-1}^1 z\cdot q(z) ~dz &= \int_{-1}^1 z\cdot \left[\sum_{k=1}^{d-1}C_k\cdot T_k(z) + \frac{1}{2}C_0\right] dz \nonumber \\
    &= \frac{1}{2}C_0 + \sum_{k=1}^{d-1}C_k \int_{-1}^1 z\cdot T_k(z) ~dz. 
    \label{eq:derivation}
\end{align}
Since Chebyshev polynomials are symmetric, we have
\begin{equation}
    \int_{-1}^1 z\cdot T_k(z) ~dz = [1+(-1)^{k+1}] \int_0^1 z\cdot T_k(z) ~dz.
    \label{eq:mean_forecast}
\end{equation}
Then Eq (\ref{eq:mean_forecast}) is zero if $k$ is even. Otherwise, according to the recurrence of Chebyshev polynomial
\begin{equation}
     \int_{-1}^1 z\cdot T_k(z) ~dz = \left[ \frac{T_{k-2}(z)}{2(k-2)} - \frac{T_{k+2}(z)}{2(k+2)} \right]\Bigg|_0^1 = \frac{2}{k^2 - 4}.
     \label{eq:int_res}
\end{equation}
Combine Eq (\ref{eq:int_res}) with Eq (\ref{eq:derivation}) and impose $\int_{-1}^1 z\cdot q(z) ~dz = \hat{x}_{t_0: T}$, we then have
\begin{equation}
    C_0 = 2\cdot \hat{x}_{t_0: T} - 4\sum_{k=1, k~\mathrm{odd}}^{d-1} \frac{C_k}{k^2 - 4}.
    \label{eq:mean}
\end{equation}

\section{Additional Results of \ref{sec:financial}}